\documentclass[pdflatex, sn-mathphys-num, iicol]{sn-jnl}
 
\usepackage{graphicx}%
\usepackage{multirow}%
\usepackage{amsmath,amssymb,amsfonts}%
\usepackage{amsthm}%
\usepackage{mathrsfs}%
\usepackage[title]{appendix}%
\usepackage{xcolor}%
\usepackage{textcomp}%
\usepackage{manyfoot}%
\usepackage{booktabs}%
\usepackage{algorithm}%
\usepackage{algorithmicx}%
\usepackage{algpseudocode}%
\usepackage{listings}%
\usepackage{subcaption}

\usepackage[thinc]{esdiff} 

\usepackage{bm} 
\usepackage{blkarray, bigstrut} 

\theoremstyle{thmstyleone}%

\newtheorem{lemma}{Lemma}

\newtheorem{remark}{Remark}%

\newcommand\blfootnote[1]{%
  \begingroup
  \renewcommand\thefootnote{}\footnote{#1}%
  \addtocounter{footnote}{-1}%
  \endgroup
}
\begin{document}

\title[Article Title]{Graph neural network for colliding particles with an application to sea ice floe modeling}


\author*[1]{\fnm{Ruibiao} \sur{Zhu}}\email{ruibiao.zhu@anu.edu.au}

\affil*[1]{\orgdiv{College of Systems and Society}, \orgname{The Australian National University}, \orgaddress{\city{ACT}, \postcode{2600}, \country{Australia}}}


\abstract{This paper introduces a novel approach to sea ice modeling using Graph Neural Networks (GNNs), utilizing the natural graph structure of sea ice, where nodes represent individual ice pieces, and edges model the physical interactions, including collisions. This concept is developed within a one-dimensional framework as a foundational step. Traditional numerical methods, while effective, are computationally intensive and less scalable. By utilizing GNNs, the proposed model, termed the Collision-captured Network (CN), integrates data assimilation (DA) techniques to effectively learn and predict sea ice dynamics under various conditions. The approach was validated using synthetic data, both with and without observed data points, and it was found that the model accelerates the simulation of trajectories without compromising accuracy. This advancement offers a more efficient tool for forecasting in marginal ice zones (MIZ) and highlights the potential of combining machine learning with data assimilation for more effective and efficient modeling.}




\maketitle

\section{Introduction}

This paper\blfootnote{This preprint has not undergone 
peer review (when applicable) or any post-submission improvements or corrections. The Version of Record of this article is 
published in \textit{Arabian Journal for Science and Engineering}, and is available online at https://doi.org/10.1007/s13369-026-11188-z.} explores the integration of Graph Neural Networks (GNNs) for modeling discrete sea ice floes in the Marginal Ice Zone (MIZ). Sea ice acts as a critical regulator of the Earth's energy balance. Sea ice's high reflectivity (albedo) plays a crucial role in moderating the global climate by reflecting solar radiation back into space \cite{climatechange_seaice, Albedo2018, Albedo2020}. However, the diminishing extent of sea ice due to warming leads to a decrease in albedo and a corresponding increase in solar energy absorption by the Earth's surface, accelerating global warming \cite{iceCO2}. Thus, the simulation of sea ice floes is a crucial component in climate modeling and prediction, playing a vital role in enhancing understanding of the Earth's climate system, especially in MIZ \cite{SeaIceModeling2020}. 

The importance of sea floe simulation in understanding and addressing the challenges posed by a changing climate cannot be overstated \cite{2019NatCC,2018WinterClimateChange,2014seaiceloss, wildfire2024}. Accurately predicted results from sea floe simulation would benefit the economy \cite{2017ArcticClimate,2020economicimpacts, hou2025green}, ecosystems \cite{ecosystem, lubitz2024climate}, astronomy \cite{meteorite2024}, and human safety \cite{2023humansafe}. The Discrete Element Method (DEM) is particularly adept at exploring the complex behavioral dynamics and spatial-temporal variations of sea ice, especially on smaller scales such as those found in the MIZ. DEM's detailed parameterizations allow for the simulation of mechanical interactions and responses to external forces like ocean drag and atmosphere drag, providing deep insights into the stress responses and movement dynamics of sea ice under various environmental conditions. Overall, DEM enhances our understanding and ability to predict sea ice behavior in these critical zones \cite{DEM2012, ignore_air2016, DEM2022, Deng2023, MontemuroDEM2023}.
One significant limitation of the DEM is its high computational demand \cite{DEM_expensive2004}, which is especially evident in the context of sea ice modeling, where simulations must cover extensive geographic areas and long-duration phenomena. Historically, constraints on high-performance computing (HPC) resources have favored continuum models, with DEM being too costly to execute \cite{SeaIceModeling2020}. This computational load primarily stems from the necessity to detect contact between neighboring elements, a process that becomes particularly costly when employing polygonal elements due to the complex algorithms required to calculate contact points \cite{DEM_expensive2022}. Consequently, this restricts the model's application over larger spatial and temporal scales. 

In general, solving forward problems typically involves fewer computational resources than solving inverse problems \cite{tarantola2005inverse, isakov2018inverse}. The GNNs provide a powerful tool for efficiently solving inverse problems by learning direct mappings from data to parameters, and GNNs have proven to be highly effective for analyzing graph-structured data, which underpins their growing application in various domains, including social networks, drug discovery, and more \cite{GCN2018, gligorijevic2021structure, merchant2023scaling}. This effectiveness arises from GNNs' ability to model relationships and interactions directly within the graph data, capturing complex connectivity patterns that are otherwise difficult with traditional neural network approaches \cite{wu2020comprehensive, zhou2020graph}.

GNNs operate on graphs by utilizing nodes and edges where the nodes represent entities and edges denote the interactions between these entities. Through techniques like message passing, nodes update their states based on the information from their neighbors, allowing GNNs to learn and generalize from the graph structure efficiently \cite{wu2020comprehensive, zhou2020graph}. This process is particularly advantageous for tasks that involve rich relational data. The motivation for applying GNNs to sea ice modeling is supported by other proven successes in similarly complex systems in other scientific domains \cite{GCN2018, gligorijevic2021structure, merchant2023scaling}. Given these capabilities, the application of GNNs to sea ice modeling presents a promising new avenue. GNNs could capture more insights by conceptualizing sea ice dynamics into graphs, where nodes represent sea ice features, including position and velocity, and edges capture interactions such as collisions. 

\begin{figure}[ht!]
\centering
 \includegraphics[width=0.8\linewidth]{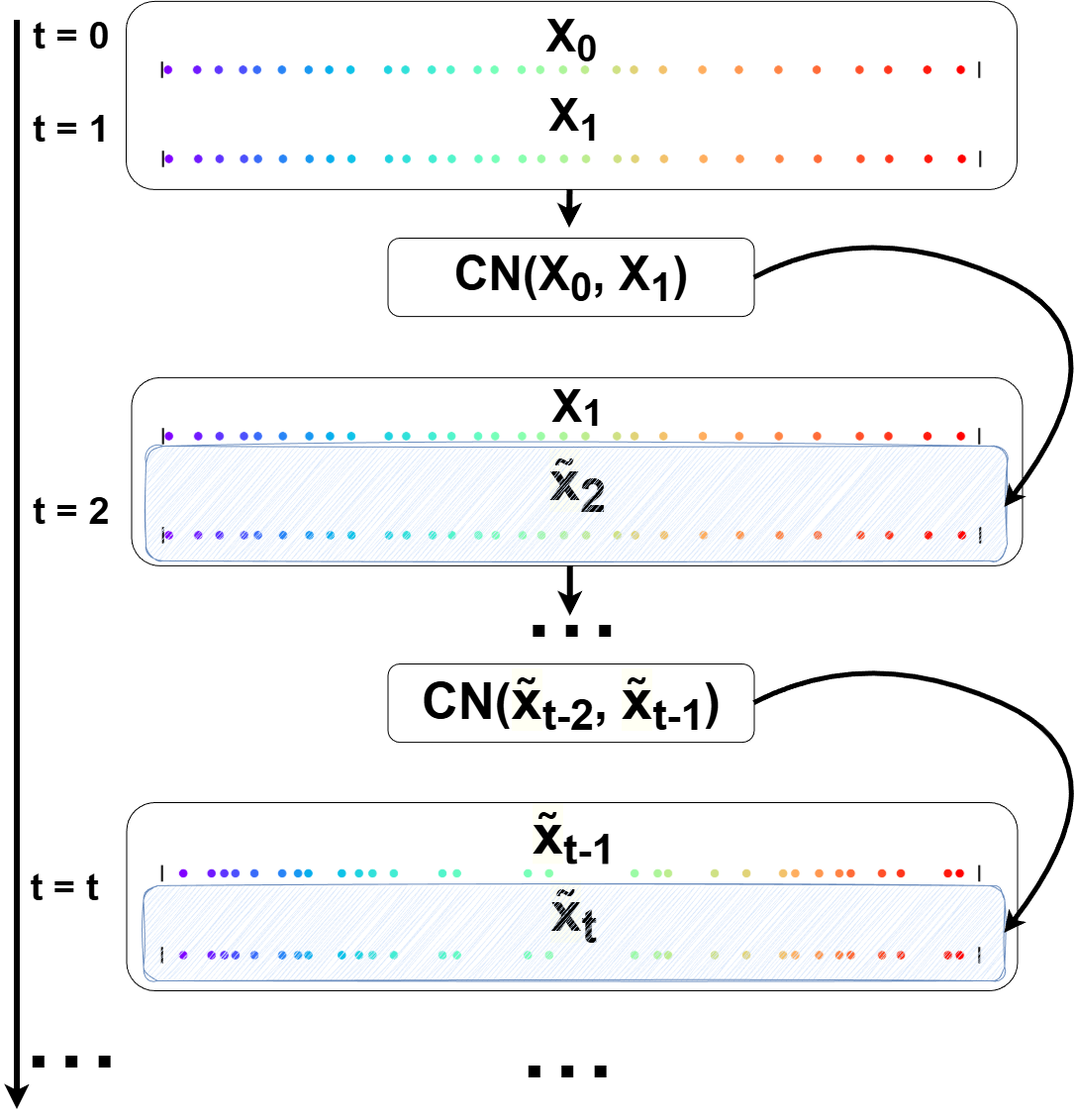}
 \caption{The illustration of CN predictions with only initial states as inputs. In the visualization, different floes are distinguished by representing each with a uniquely colored circle. The predicted state at time \( t \) is denoted with a tilde (\(\tilde{X}_{t}\)) over the head, while the ground truth state at time \( t \) is represented without the tilde (\(X_{t}\)). The white box represents the input of the model, and blue shaded box stand for the output of the model, and the arrow shows the data flows. The proposed model CN recursively utilizes information from the previous two-time steps to forecast the state at the subsequent time step, predicting the states for all times, except for the initial two-time steps which serve as ground truth inputs. }
 \label{fig:flowchart}

\end{figure}

Previous efforts to simulate sea ice dynamics within the MIZ have predominantly employed Discrete Element Methods (DEM), focusing on the physical interactions of individual ice floes. These methods, while detailed, are computationally intensive and often struggle to scale up due to their high computational demands \cite{DEM_expensive2004, SeaIceModeling2020}. Surprisingly, despite the rapid advancement in machine learning technologies \cite{liu2025mathematical, hou2025machine, xie2024enhanced, liu2024detracking}, there has been a noticeable absence of research exploring the use of neural networks for sea ice simulation in the MIZ. This study contributes to this area by introducing a GNN-based sea ice simulation model: proposed Collision-captured Network (CN) as Figure \ref{fig:flowchart} coupled with data assimilation (DA) techniques, inspired by the Interaction Network \cite{battaglia2016interaction}. By leveraging the inherent graph structure of sea ice interactions in the MIZ, the proposed model offers an innovative solution that reduces computational overhead while enhancing simulation fidelity, providing a promising new direction for future climate modeling efforts.
\\

\section{Methods}
\label{section:Methods}

\subsection*{Problem setting}

This study employs a one-dimensional sea ice simulation as the primary dataset to facilitate a more focused analysis of the collision mechanisms between different sea ice floes. This simplification is strategically chosen to reduce the computational complexity associated with multidimensional simulations while retaining the essential dynamics of sea-ice interactions. By constraining the simulation to one dimension, the critical factors that influence floe collisions, such as velocity and position can be isolated and more effectively captured. 

In a one-dimensional setting, individual floes do not undergo rotation, eliminating the presence of tangential forces at points of contact. Additionally, external forces such as ocean-induced and atmospheric drag are disregarded. This simplification focuses the study on elucidating the mechanisms of collision. 

Also, disks are used to represent floes in the sea ice modeling to balance geometric simplicity and physical accuracy, enhancing the simulation's effectiveness. The disk shape simplifies the computational process, facilitating more straightforward calculations for interactions such as collisions and contacts, which is essential in scenarios with numerous floes or intricate environmental interactions \cite{disk1999}. This form also provides a realistic representation of real ice floe behaviors under various conditions, as evidenced in simulations that analyze the impact of pancake ice floes on cylindrical structures \cite{disk2012}. Thus, the choice of using disks as a representation for floes strategically balances the need for detailed physical representation with computational efficiency, which proves vital in comprehensive simulations of ice dynamics and interactions. 

Currently, there is a lack of benchmark datasets specifically designed for training and testing neural network models in the context of sea ice simulation. For generating the ground truth data, the DEM and some governing equations \cite{ApplicationDEM} are used as shown in the following sections, and the data with visualization shown in Supplementary information \ref{appendix:Visualization} is validated. For example, a floe cannot pass another floe and jump onto the other side of the floe in the one-dimensional simulation setting. Assuming denoting floes from left to right sequentially (Fig.\ref{fig:floe_setup}), the system would maintain $x^i < x^j$, where $i<j$ and $x^i$ is the x-axis position of floe $i$ at the same time. 

\subsubsection*{The equations of motion for sea floes}
Let $x^i$ and $v^i$ represent the center position and velocity of the $i$-th floe, and linear velocities for sea ice floe $i$ are defined as following:
\begin{equation} \label{eq:velocities}
\begin{split}
v^i & = \diff{x^i}{t} 
\end{split}
\end{equation}
\begin{itemize}
\item $v^i$ is linear velocity
\item $x^i$ is the position
\end{itemize}

The right direction is defined as positive and the left direction as negative for force, acceleration, and velocity. Thus, $x^i$ and $v^i$ are treated as scalars in the plane. 

\subsubsection*{Translational momentum balance}

The translational momentum balance based on Newton's laws of motion $F = ma$ for an ice floe $i$ is 
\begin{equation} \label{eq:translation2}
\begin{split}
m^{i} \diff[2]{x^{i}}{t} & = \sum_{j}f_{n}^{ij}
\end{split}
\end{equation} where $x^i$ is the center position of the floe $i$, and $m^i$ is the mass of floe $i$. When floe $j$ contacts with floe $i$, $f_{n}^{ij}$ is contact-normal force between floe $i$ and floe $j$.

\subsubsection*{Contact-normal force with hookean linear elasticity}

\begin{figure}
 \centering
 \includegraphics[width=0.8\linewidth]{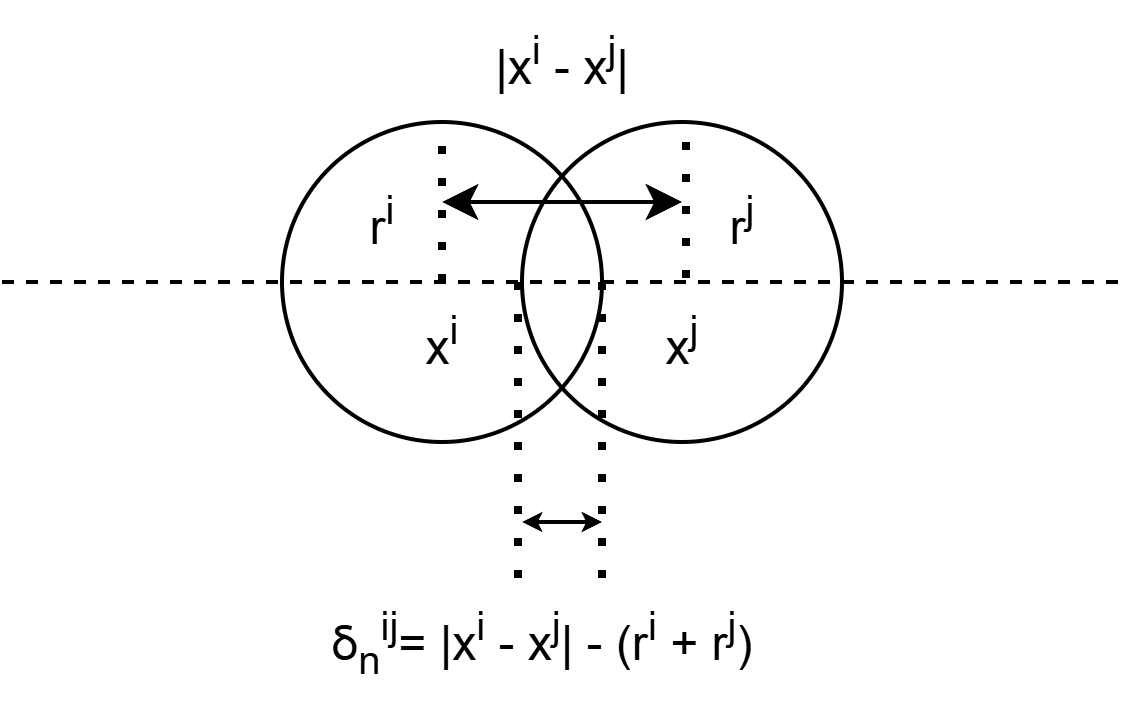}
 \captionof{figure}{The illustration of computing the overlap distance $ \delta_n^{ij}$ when the two floes contact with each other.}
 \label{fig:floe_interaction}
   
\end{figure}

The stresses resulting from axial compressive strain can be modeled as nonlinear \cite{nonlinearElasity2013}. However, a nonlinear model requires extremely small time steps due to the numerical instability. Thus, a common approach for capturing the jamming behavior of the interaction of two floes in sea ice simulation is adopted \cite{Hookean2011, ignore_air2016, ApplicationDEM}. The resistive force to contact compression, based on overlap distance $ \delta_n^{ij}$ between floe $i$ and floe $j$ (Fig.\ref{fig:floe_interaction}) rather than complete elastic collision, is computed by Hookean linear elasticity \cite{Hookean1979, Hookean2011, Hookean2016}. This contact force is non-zero when there is an overlapping area between two floes. Otherwise, the contact force is zero. This is a common method in discrete element simulations to model the contact between two different cylindrical ice ﬂoes \cite{ApplicationDEM, Hookean1979,ergenzinger2011discrete, luding2008introduction}.

When $\delta_n^{ij} = |x^i - x^j| - (r^i + r^j) < 0$, \newline
\begin{equation} \label{eq:contact}
\begin{split}
f_n^{ij} & = A^{ij}E^{ij}\delta_n^{ij}\\
& = R^{ij}\text{min}(h^i,h^j)E^{ij}\delta_n^{ij} \\
& = \frac{2r^ir^j}{r^i+r^j} \cdot \text{min}(h^i,h^j)E^{ij}\delta_n^{ij}
\end{split}
\end{equation}

where $R^{ij} = \frac{2r^ir^j}{r^i+r^j}$ is the harmonic mean of floe $i$ radius $r^i$ and floe $j$ radius $r^j$. $E^{ij}$ is the Young's modulus, and $r^i$ and $r^j$, $h^i$ and $h^j$ are ice floe $i$ radius and ice floe $r$ radius, ice floe $i$ thickness and ice floe $j$ thickness, respectively \cite{damsgaard2021effects, ApplicationDEM}. 

\subsubsection*{Experiment setup}

In the simulations, unitless variables for all parameters and state variables are employed to enhance the generalizability and applicability of the model across various scenarios. This decision allows the model to adapt to different scales and environments without needing specific unit conversions. Also, it is essential to set the initial velocities of the floes to relatively high values and to constrain their positional domain. These parameters are crucial to ensure that collisions occur frequently enough to provide substantial training data (Fig.\ref{Fig:30nodes_dataset_snapshot} and Fig.\ref{Fig:10nodes_dataset_snapshot}), and to facilitate the verification of compliance with physical laws. High initial velocities increase the likelihood and frequency of collisions within a controlled spatial domain, thereby enriching the dataset with diverse interaction scenarios. Additionally, by limiting the extent of the positional domain, the accuracy and realism of the floe dynamics can be more easily monitored and evaluated, ensuring that the simulation remains within acceptable physical bounds. 

The Euler method is used in equation \ref{eq:Euler1D} combining mentioned governing equations \ref{eq:velocities}, \ref{eq:translation2}, and \ref{eq:contact} to march into the next time step. In this case, $v_{t_{j}}^i$ is the velocity of $i$ at time $t_{j}$, $x_{t_{j}}^i$ is the x-axis position of $i$ at time $t_{j}$, and $f_{t_{j}}^i$ is the contact force of $i$ at time $t_{j}$ if there is any. 
\begin{equation} \label{eq:Euler1D}
\begin{split}
v_{t_{j}}^i & = v_{t_{j-1}}^i + \frac{f_{t_{j-1}}^i}{m^i}dt \\
x_{t_{j}}^i & = x_{t_{j-1}}^i + v_{t_{j}}^idt
\end{split}
\end{equation}

\begin{figure}
 \centering
 \includegraphics[width=1\linewidth]{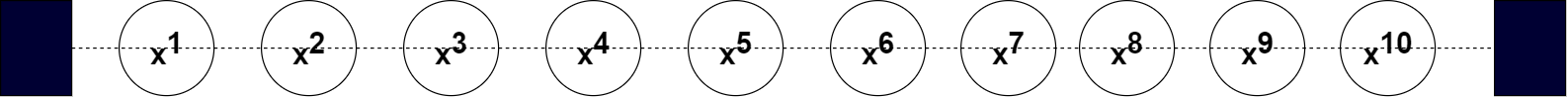}
 \captionof{figure}{\textbf{The illustration of experiment settings}. The boundaries are shown as black walls, and the floes are represented as circles.}
 \label{fig:floe_setup}
   
\end{figure}

The initial conditions for one dimension setting are the following: the radius of each floe is 1; the thickness of each floe is 1; the position of each floe is randomly generated in the domain, the initial velocity is from 150 to 200, and the direction of velocity can be left as negative or right as positive along with x-axis, which can be interpret as approximately 150 to 200 meter per hour if compared to the real observation. This is on the same order of magnitude as observed floe drift in real sea-ice fields in certain areas \cite{lund2018arctic}. The larger number of unitless velocities can make floes traverse the bounded domain more quickly and are more likely to collide with each other or with the domain boundaries, leading to more frequent collision events. This is desirable for a collision-capturing training scenario, as it provides a richer dataset of collision outcomes for model learning. Also, the boundary is 0 and 100, and 0 and 200 for 10 and 30 nodes simulation, respectively. Furthermore, sea floes perform the same collision mechanical interactions with the boundary as they contact with other floes, except that the boundaries are stationary (Fig.\ref{fig:floe_setup}). 

The time step length and Young's modulus are carefully selected to accurately simulate the dynamics of sea ice floes. The need for smaller time steps in complex models arises from the necessity to accurately capture the dynamic behaviors and interactions within the collision system. Smaller time steps ensure that the numerical methods can adequately resolve these interactions without missing critical dynamics \cite{j2017analytical, selectTimeStep2004}. For example, larger time steps, such as $dt = 10^{-2}$ and $dt = 10^{-3}$, have been tested in the data generation setting, and they cannot work since the overlapping area of floes would be larger. This would cause the contact force to be significantly large based on the formula, and the updated x-axis position can be pretty inaccurate based on the extremely large acceleration. For example, a floe can pass to the other side of its neighbor due to the significantly large contact force, which is impossible in this physical domain. Thus, the model is adjusted to smaller time steps such as $dt = 10^{-4}$ for the data generation. For Young’s modulus $E$, reducing the elastic stiffness is a known technique in DEM simulations to permit larger stable time steps and speed up computations \cite{damsgaard2021effects, tuhkuri2018review, ApplicationDEM}. Using a very high Young’s modulus $E$ would thus yield extremely stiff contacts and large, rapidly varying contact forces, necessitating extremely small time-step sizes to maintain numerical stability. In contrast, a lower Young’s modulus $E$ softens the contacts, which caps the peak contact force and makes collisions less abrupt. Thus, $E = 2 \times 10^7$ was chosen in this study to reach a reasonable balance between computational efficiency and elastic compressibility \cite{ApplicationDEM}, although this Young’s modulus is lower than the observed value \cite{E1999,E2003}.

By optimizing these parameters, the simulations realistically capture the mechanical properties and interactions of the ice, and aid in effectively training and validating neural network models.

\subsection*{Related models}
Sea-ice floe dynamics with collisions pose a learning problem that when contacts occur, the state can change abruptly, with post-collision velocities depending on the coupled states of the colliding bodies and contact parameters. This makes the data distribution strongly imbalanced—dominated by no-contact steps, while the important behavior is concentrated in rare, high-impact events. A closely related collision benchmark in the Interaction Network (IN) highlights exactly this difficulty. For more than $99\%$ of steps, a ball is not in contact and its next-step velocity equals its current velocity, whereas the remaining steps require learning a complex, state-dependent collision response \cite{battaglia2016interaction}. Hamiltonian Neural Network (HNN), which represents dynamics through differentiable equations and recovers time evolution by taking gradients, is a typical example of a physics-learning machine learning approach that can struggle in such settings, since HNN requires differential equations to successfully describe the whole system within the space and time domains, and these settings cannot satisfy the requirement of HNN \cite{greydanus2019hamiltonian}. 

The Interaction Network (IN) treats objects as nodes and relations as edges, and the Graph Network-based Simulator (GNS) extends IN with an encoder-decoder mechanism, noise injection, and stacked graph layers \cite{battaglia2016interaction, partical}. These models perform well in predicting particle movement trajectories. These two models can be directly aligned with sea-ice floe modeling, and these models are tested for comparison.

\subsection*{The CN model framework}
The total time $T$ can be discretized into $n$ time steps with $t_j = j\frac{T}{n}, j \in [0, \dots,n-1]$ .
The state of floes at time \( t_{j} \) is denoted as $X_{t_{j}}$. The position and velocity at time \( t_{j} \) are denoted as $x_{t_{j}}$ and $v_{t_{j}}$, respectively. $X_{t_{j-2}:t_{j-1}} = [x_{t_{j-2}}, x_{t_{j-1}}, v_{t_{j-1}}, r]$ is previous consecutive objects' information that contains the second last predicted position $x_{t_{j-2}}$, last predicted position $x_{t_{j-1}}$, the most recent inferred velocity $v_{t_{j-1}}$ and radius $r$. 

\begin{figure}
 \centering
 \includegraphics[width=1\linewidth]{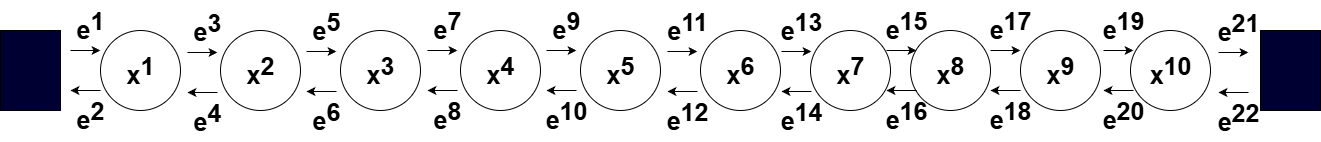}
 \captionof{figure}{The illustration of graph construction for GNN. The boundaries are shown as black walls, and $e$ represents the edge, and $x$ stands for the node.}
 \label{fig:floe_edge_setup}
   
\end{figure}

When constructing the graph, the floes and two boundaries can be treated as nodes, and the bidirectional edges are constructed between nodes and their neighbors (including the boundaries) (Fig.\ref{fig:floe_edge_setup}). The edge set is defined by neighbor adjacency rather than distance-based thresholds: edges are created only between designated neighboring nodes, irrespective of their instantaneous separation. With this construction, the graph connectivity remains constant over time, and therefore, the number of edges is fixed throughout the simulation. This fixed relational structure provides two practical benefits: it removes the computational overhead and potential instability associated with dynamically rebuilding graphs at every time step, and it ensures that the learned interaction functions operate on a consistent topology, so that changes in predicted dynamics arise from evolving node and edge features rather than from time-varying graph structure. The edge feature $E_{t_{j}} = [e_{ij}; e_{ji}]$ is the concatenation of the displacement of floe $i$ to floe $j$ and the displacement of floe $j$ to floe $i$. 

To identify which node is the sender or the starting point for each directed edge in the graph, $R_s$ is defined. The entries of $R_s$ are binary with $R_s[i, j] = 1$ if the $j$-th edge originates from node $i$ and $R_s[i, j] = 0$ if the $j$-th edge does not originate from node $i$. Meanwhile, to identify which node is the receiver or the endpoint for each directed edge in the graph, $R_r$ is defined. The entries of $R_r$ are binary with $R_r[i, j] = 1$ if the $j$-th edge terminates at node $i$, and $R_r[i, j] = 0$ if the $j$-th edge does not terminate at node $i$. More details regarding relation and permutation matrices are in Supplementary information \ref{appendix:Relation_Matrices}.

The proposed Collision-captured Network (CN) has been defined as, 
\small
\begin{equation} \label{CN_1D}
\begin{split}
E_f
= \phi_{\theta_1} ([R_r^TX_{t_{j-2}:t_{j-1}}; R_s^TX_{t_{j-2}:t_{j-1}};E_{t_{j-1}}])\\
X_{t_{j}} 
= \gamma_{_{\theta_2}}([{X}_{t_{j-2}:t_{j-1}}; R_rE_f])
\end{split}
\end{equation}
\normalsize



Since the message passing in GNN is similar to multi-layer perception (MLP), $\phi_{\theta_1}$ with $\theta_1$ learnable parameters and $\gamma_{_{\theta_2}}$ with $\theta_2$ learnable parameters are set to be MLP with four 150-length hidden layers and one 100-length hidden layer, respectively \cite{message_passing1990}. When the simulation has $N$ floes and 2 boundaries, the number of edges is $N_e = 2(N+1)$. Thus, both $R_r$ and $R_s$ are $(N+2) \times N_e$ matrix. $E_{t_{j-1}}$ is $N_e$-length vector. $X_{t_{j-2}:t_{j-1}}$ is $(N+2) \times 4$ matrix, since it contains 4 different features as mentioned before within it. The $\phi_{\theta_1}$ serve as the aggregating function \cite{GNN2005, GNN2009} the edge effects by taking the concatenation of interaction terms and edge features. The output of $\phi_{\theta_1}$ is $(N+2) \times D_{\phi}$ matrix, where the $D_{\phi}$ is the output dimension of the $\phi$. Thus, $R_rE_f$ is the aggregation edge effects working on the nodes with $(N+2) \times N_e$ dimension. Furthermore, $\gamma_{_{\theta_2}}$ serve as updating function \cite{GNN2005, GNN2009} in GNN by taking the input of the concatenation of node features and aggregated edge effects working on the nodes.

The proposed model differs from the standard IN formulation in three deliberate and practically important ways: (i) the state representation used for prediction, (ii) the edge features to better encode collision-relevant geometry, (iii) the activation function.


A standard Interaction Network predicts the next-step state by conditioning on the full state at the most recent time step, typically including both position and velocity as node features. This design is appropriate when velocities are directly available from a simulator or can be reliably measured. In contrast, our proposed model is intended to align with more realistic observation conditions in sea-ice applications, where position is the primary quantity that can be consistently observed, while velocity is often unavailable or substantially noisier. Accordingly, we do not require velocity as an explicit input. Instead, we use positions from the two most recent time steps to derive the most recent-step velocity through a finite-difference approximation. This formulation retains essential dynamical information while reducing reliance on velocity observations. 


A second distinction is the explicit incorporation of collision-relevant geometry into edge features. In standard Interaction Networks, interactions are only represented via learned message functions, but edge features are not necessarily constructed. Our model implements the edge representation with a specialized feature: the instantaneous inter-floe displacement. This displacement signal directly encodes proximity, which is a primary determinant of contact and collision behavior. By injecting the instantaneous inter-floe as edge features, the model becomes more sensitive to contact configurations and is better positioned to learn physically plausible responses, such as resolving collision and preventing unrealistic overlap.


All activation functions are Mish functions $Mish(x)$ \cite{misra2019mish} as equation \ref{eq:mish} to improve performance, as compared to all ReLU functions \cite{ReLU2000} in IN. This Mish activation function can improve the performance of the model since this function is non-monotonic, non-monotonic derivative, not saturated, and it has infinite continuity and an approximating identity near the origin. Unlike the ReLU, the non-monotonic utilizes the negative derivative and improves expressiveness. Compared to the sigmoid function, Mish is not saturated, and it avoids near-zero gradients to make the model learn more efficiently and effectively \cite{misra2019mish}. 

The model's output can be velocity or position at time $t_{j}$, proving the velocity would be better based on the error analysis in Supplementary information \ref{appendix:Error_analysis}. Thus, the loss function for training is:

\begin{equation} \label{eq:loss}
\mathcal{L} = \frac{1}{n}\sum_{j=2}^{n}(\text{CN}(X_{t_{j-2}:t_{j-1}})-v_{t_{j}})^2 
\end{equation}

 To train the proposed model, the training pairs ($X_{t_{j-2}:t_{j-1}}, v_{t_{j}}$) are randomly sampled, and be computed above loss function between the predicted velocity and ground truth velocity of each floe. Based on Equation \ref{eq:loss}, the optimization problem is to minimize the loss with $\theta_1$ and $\theta_2$. Then, the parameters of the model are optimized over this loss with the Adam optimizer for its optimization performance in accelerating convergence \cite{kingma2014adam}. More details about training are shown in the Supplementary information \ref{appendix:Training}. A key feature of the proposed model is its parameter-sharing structure, which enables the model to generalize effectively while maintaining statistical efficiency. Parameter sharing means that every edge message is computed using the same $\phi_{\theta_1}$, and every node state is updated using the same $\gamma_{\theta_2}$ across all time steps, rather than learning separate functions for particular floes or particular interacting pairs, unlike the GNS that employs deep stacks of message-passing blocks, each layer can have its own learnable weights, even though the layer structure is repeated. This principle aligns with the core motivation of IN style models: sharing the relation and object reasoning mechanisms enables the learned dynamics to generalize across different numbers and configurations of entities, because the model learns reusable interaction rules rather than instance-specific behavior \cite{battaglia2016interaction}. By sharing parameters across tasks or components within the model, the overall number of learnable parameters is reduced, thus mitigating overfitting risks and reducing training time. This approach leverages the benefits of shared information, where parameters trained in one context can transfer knowledge to another, enhancing the model's adaptability across varying data distributions. Consequently, parameter sharing allows the model to make efficient use of available data, resulting in faster convergence and lower computational costs, without compromising predictive accuracy \cite{pham2018efficient}.

In order to initialize the proposed model in prediction, it is imperative that the initial states for the first two consecutive time steps, which are $t_1$ and $t_2$, along with the relevant edge information $R_r$, be fed into the model as ground truth. This foundational data acts as a crucial starting point, enabling the model to set initial conditions accurately. Once initialized, the model leverages these inputs to predict the object states in the next step. To facilitate subsequent predictions, the model is designed to recursively incorporate its own predictions as inputs. This recursive application allows the model to continually refine its predictions based on the evolving dynamics captured by the updated object states. Thus, the model functions in a feedback loop, where the output from one step serves as the input for the next, enhancing the model's predictive capability over time.

Furthermore, observational data is combined with the proposed model forecasts to refine predictions by using data assimilation. Specifically, the Ensemble Kalman filter (EnKF) and the Ensemble Transform Kalman filter (ETKF) are used to better handle the uncertainties in the system in the long-time trajectories. More details about both Kalman filters are in the Supplementary information \ref{appendix:KF}.

\section{Results}

The approach was verified using unseen truth data. The results show that CN not only accelerates the simulation of sea ice trajectories but also generalizes well to much longer time scales than those included in the training set, predicting sea ice dynamics over extended periods with high accuracy. This improvement in efficiency, combined with the model's ability to predict accurately over longer time scales, makes the proposed approach a promising tool for more effective and computationally efficient sea ice modeling, particularly for applications in MIZ forecasts.

\subsection*{Model performance}

Three root mean squared error (RMSE equation from \ref{eq:RMSE}) metrics are reported to separate short-horizon accuracy from long-horizon simulation stability. One-step position RMSE is defined as the root mean squared error of the average next-step position prediction when the model is conditioned on ground-truth inputs. Similarly, one-step velocity RMSE is the RMSE of the average next-step velocity prediction under ground-truth inputs. Also, simulation RMSE measures long-horizon error: it is computed as the time-average of position RMSE over the entire simulation when the model is input only with the initial ground-truth states and then evolves autoregressively. In addition to RMSE, the pattern correlation coefficient (PCC equation from \ref{eq:PCC}) is used to quantify how well the predicted trajectories preserve the ground-truth spatiotemporal pattern; as with standard correlation measures, PCC takes values in $[-1,1]$, where negative values indicate an inverse (anticorrelated) relationship between predicted and target patterns.

Table \ref{table:model_performance_comparision} shows that all CN variants outperform the Interaction Network (IN) baseline. This improvement is consistent with two design choices. First, CN explicitly augments the interaction model with specialized edge features (including direct inter-floe displacement information), which facilitate the message-passing mechanism to preserve separation and maintain physically plausible spacing during close-contact events. Second, CN conditions its updates on two consecutive position states, using these to infer the most recent velocity internally, rather than requiring externally provided velocities. By contrast, the original IN formulation predicts dynamics via shared relation- and object-centric functions applied over a graph, but does not intrinsically enforce domain-specific edge geometry. Empirically, the fact that CN with one step already improves over IN indicates that edge-feature implementation and replace ReLU with Mish activation function can shift performance from moderate PCC to high PCC, supporting the interpretation that explicit displacement-aware relational features are particularly beneficial in floes simulation tasks.

Across CN variants, Table \ref{table:model_performance_comparision} indicates that Mish yields slightly better performance than SiLU, while both outperform ReLU. This aligns with prior findings that smoother, self-gated nonlinearities can improve optimization and representational fidelity in deep networks. Since Mish is explicitly related to the Swish family, the similarity in performance between Mish and SiLU is expected, whereas ReLU can be slightly less expressive \cite{hendrycks2016gaussian, ramachandran2017searching, misra2019mish}.

The Graph Network-based Simulator (GNS) performs poorly in Table \ref{table:model_performance_comparision}, with negative PCC values indicating that the learned simulation captures an oppositely varying pattern relative to the target. While GNS is a strong general-purpose learned simulator that evolves particle systems via stacked message-passing blocks, long-horizon performance is known to depend sensitively on architectural depth and the handling of error accumulation. In collision floe dynamics, interactions are fundamentally local (short-range proximity), and propagating information beyond immediate neighbors can be unnecessary or even harmful. In message-passing GNNs, $k$ rounds of propagation yield representations that increasingly mix information over the $k$-hop neighborhood, and deeper stacks are therefore more prone to blending local identities \cite{nikolentzos2020k}. This connects to the well-documented over-smoothing phenomenon in deep GNNs, where node embeddings become indistinguishable as depth increases, degrading performance on tasks that require sharp local distinctions \cite{wu2023demystifying, kelesis2025analyzing}. Additionally, GNS  uses a five-time-step information to forecast the next state; however, in discrete time, a two-step position history is already adequate to derive the velocity or position via finite differences \cite{DEM2012,DEM_expensive2004, DEM_expensive2022}. In this sense, conditioning on longer histories may introduce redundant or noisy temporal correlations that are not required by the underlying DEM-generated dynamics, potentially contributing to degraded PCC and inflated RMSE in simulation.

The results from this study on sea ice simulation, explicitly focusing on one-dimensional scenarios involving 10 and 30 floes collisions, demonstrate significant achievements in the predictive accuracy of this newly developed model. The model's effectiveness is underscored by high Pattern Correlation Coefficients (PCC equation from \ref{eq:PCC}) and low Root Mean Squared Errors (RMSE equation from \ref{eq:RMSE}) across both scenarios. Notably, all variables in the proposed model are unitless to ensure the generalizability of results across different scenarios and settings (More details in Supplementary information \ref{appendix:Skill_Scores}). Visualization can be checked in Supplementary information \ref{appendix:Visualization} including snapshots for the prediction (Fig.\ref{Fig:10nodes_snapshot} and Fig.\ref{Fig:30nodes_snapshot}). Further details about methodologies are comprehensively discussed in the Methods section. 

\begin{table*}[ht]
\caption{Model performance comparison}
\label{table:model_performance_comparision}
\small
\setlength{\tabcolsep}{4pt} 

\begin{tabular*}{\linewidth}{ccccc} \toprule
{Model} & {Simulation RMSE} & {one-step position RMSE} & {one-step velocity RMSE} & {PCC} \\ 
\midrule
\text{GNS \cite{partical}} & 8.897476 & 0.001089 & 1.536386 & -0.243754\\
\text{Interaction Network \cite{battaglia2016interaction}} & 11.490381 & 0.000038 & 0.373804 & 0.567590\\  
\text{CN with one step} & 4.482929 & 0.000010 & 0.070539 & 0.712879\\
\text{CN with Mish} & 2.334024 & 0.000006 & 0.028410 & 0.934317\\ 
\text{CN with ReLU} & 3.048421 & 0.000015 & 0.117186 & 0.894427\\ 
\text{CN with SiLU} & 2.452933 & 0.000009 & 0.065283 & 0.927664\\
\bottomrule
\end{tabular*}

\end{table*}

\begin{figure*}[t!]
    \centering
    \begin{subfigure}[t]{0.45\textwidth}
        \centering
        \includegraphics[width= 1\linewidth]{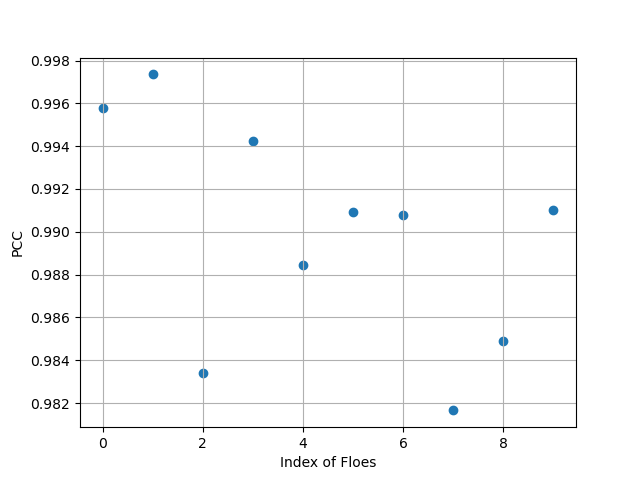} 
        \caption{PCC for 10 floes situation with the average of 98.98\%}\label{Fig:PCC_10ndoes}
    \end{subfigure}%
    ~ 
    \begin{subfigure}[t]{0.45\textwidth}
        \centering
        \includegraphics[width= 1\linewidth]{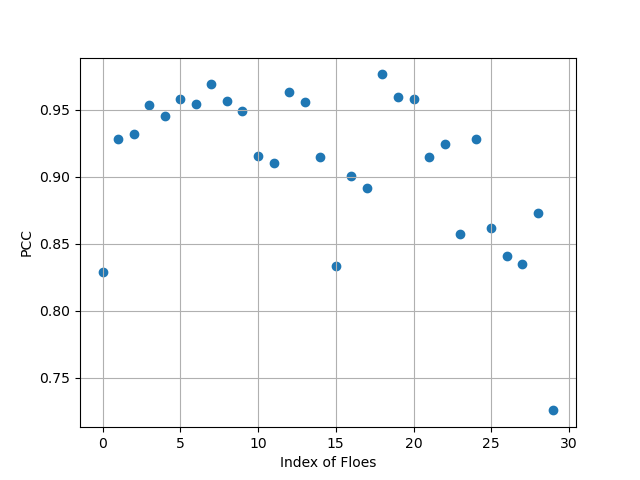}
        \caption{PCC for 30 floes situation with the average of 91.06\%}\label{Fig:PCC_30ndoes}
    \end{subfigure}
    \caption{The PCC plots for CN in inference. The x-axis is the index of floes, and the y-axis is the PCC value.}\label{Fig:PCC}
 \noindent\rule{1\linewidth}{1pt}\\
\end{figure*}

For the system with 10 floes, the PCC was high at 98.98\% (Fig.\ref{Fig:PCC_10ndoes} and Table \ref{table:pcc_10nodes}), indicating an almost perfect alignment with the observed trends, suggesting that the model can accurately simulate the interactions and dynamics of a smaller number of ice floes. Moreover, the RMSE for this scenario was 1.16, which is relatively low considering the domain range of 100. This indicates that the average deviation of the simulated values from the actual data is minimal, at only about 1.16, translating to an average deviation per floe of approximately 0.12.

In the more complex scenario involving 30 floes, the model also performed with a PCC of 91.06\% (Fig.\ref{Fig:PCC_30ndoes}), reinforcing its capability to maintain high accuracy even as the number of interactions increases. The RMSE for this setup was 3.01, set against a domain range of 200. Similar to the 10-floe scenario, this error metric suggests that the model maintains a high degree of precision. The average deviation across all 30 floes is about 3.01, with an individual floe deviating by only about 0.10 from the ground truth.

Additionally, the visualizations from the simulations confirm that the model adheres to physical rules, showcasing trajectories where sea ice floes do not pass through each other or cross defined boundary limits. This aspect of the simulation is critical as it reflects the model's ability to not only predict positions and movements accurately but also to ensure that these predictions respect the spatial constraints and interactions dictated by physical laws.

\subsection*{Generalization ability}

To assess generalization beyond the training regime, we compared the simulation of the model against trajectories generated by DEM as ground truth. Generalization was quantified using PCC and RMSE. In climate-model evaluation studies, PCC values exceeding 0.8 are commonly interpreted as indicating strong replication of observed spatial patterns \cite{salazar2024cmip6, christophersen2024oceanic}. Under this criterion, the proposed model maintains high accuracy in generating a 20,000 time steps simulation, achieving a PCC of 0.871 and a simulation RMSE of 2.334, which indicates that the learned simulation preserves the dominant spatiotemporal structure of the DEM trajectories over a substantially extended horizon.

\begin{table}[ht]
\caption{Model generalization comparison}
\label{table:model_generalization_comparision}

\begin{tabular*}{\linewidth}{ccc} \toprule
{Simulation time range} & {PCC} & {Simulation RMSE}  \\ 
\midrule
10000 & 0.934 & 2.334\\ 
20000 & 0.871 & 4.256\\ 
30000 & 0.791 & 7.230\\ 
40000 & 0.576 & 11.055\\ \bottomrule

\end{tabular*}

\end{table}

As the simulation range increases shown in Table \ref{table:model_generalization_comparision}, PCC declines to 0.791 in generating 20,000 time steps simulation and further to 0.576 in generating 20,000 time steps simulation, while simulation RMSE increases to 7.230 and 11.055, respectively. These values indicate error accumulation and reduced pattern fidelity with the increase of simulation range. The generalization capability of the proposed newly developed model for simulating sea ice dynamics demonstrates remarkable performance, particularly in its ability to predict beyond the scope of its training data. Despite the training dataset comprising only 10,000-time steps, the model successfully projects plausible trajectories for up to 20,000 steps in simulations involving both 10 and 30 floes (Fig.\ref{Fig:30nodes_generalization_snapshot} and Fig.\ref{Fig:10nodes_generalization_snapshot}). This capacity not only highlights the robustness of the model, but also its high fidelity in long-term prediction scenarios.

The model's ability to adhere to physical rules over these extended simulations is critical. It ensures that the trajectories of the sea ice floes do not unrealistically intersect or breach predefined boundaries, thus maintaining realistic simulations over time. By proving to be capable of reliable predictions over durations much longer than the training period, the model shows its potential to deliver insights and forecasts for unseen situations. 

\subsection*{Efficient inference}

In the experiments, the performance of the proposed model is compared against a traditional model, which is DEM used for generating the ground truth, in simulations involving varying numbers of sea ice floes. The simulations were conducted first with 10 floes and then with 30 floes, recording the total execution time for each scenario.
The results were highly favorable for the proposed model. When the simulation involved 10 floes, the running time for the proposed model was only 7.3 seconds compared to 7.7 seconds for the traditional model. Notably, as the complexity increased to 30 floes, the proposed model exhibited a minimal increase in running time, clocking in at 8.9 seconds. In contrast, the traditional model's running time ballooned to 24 seconds under the same conditions. 

The proposed model demonstrated significant performance advantages over the traditional model in simulations with different numbers of floes. When tested with 10 floes, the improvement was modest, outperforming the traditional model by 5\%. This slight edge can be attributed to the overhead involved in initializing and setting up the GPUs, although minimal, which affects performance when the computational scale is small \cite{dubey2019gpu}. However, as the complexity of the simulation increased to 30 floes, the benefits of the proposed model became more pronounced, with the proposed model achieving a 63\% improvement in performance over the traditional model on the hardware configuration of the Intel i7-11800H CPU and NVIDIA RTX 3080 Laptop GPU. The DEM baseline is executed on CPU NumPy (Numerical Python) \cite{harris2020array}, while the proposed method is accelerated by GPU and PyTorch \cite{paszke2019pytorch, pytorch}. To assess performance on a more recent platform, the 30 floes case is also tested using the Intel Core i9-13900K CPU and an NVIDIA RTX 4090 GPU. Under this configuration, the proposed method requires 3.9 seconds on average, whereas the traditional DEM baseline requires 20 seconds. This difference indicates the GNN-based model's ability to handle more complex interactions efficiently, making the proposed model particularly effective in scaled-up settings. However, the proposed model requires about 16 hours of training on a single NVIDIA RTX 4090 GPU to achieve high performance in simulation.


\begin{figure*}[t!]
    \centering
    \begin{subfigure}[t]{0.45\textwidth}
        \centering
        \includegraphics[width= 1\linewidth]{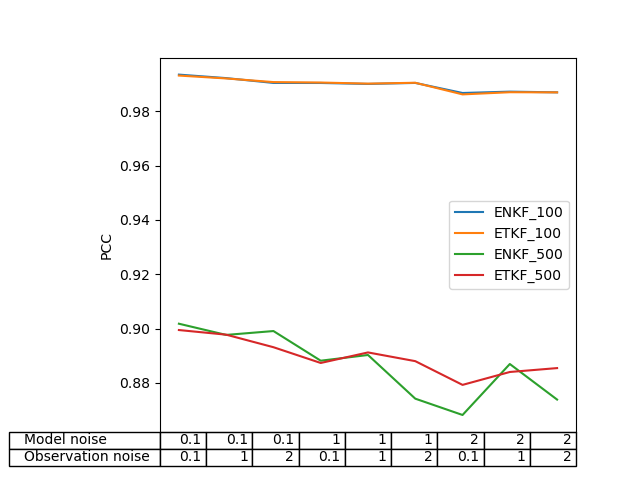} 
        \caption{PCC for 10 floes situation with ENKF and ETKF}\label{Fig:DA_PCC_10ndoes}
    \end{subfigure}%
    ~ 
    \begin{subfigure}[t]{0.45\textwidth}
        \centering
        \includegraphics[width= 1\linewidth]{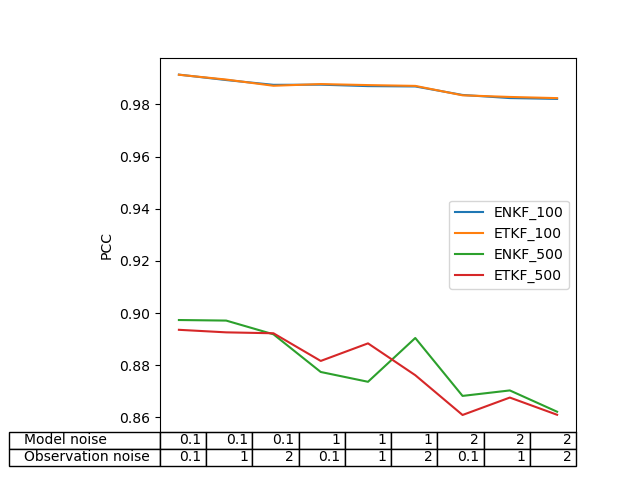}
        \caption{PCC for 30 floes situation with ENKF and ETKF}\label{Fig:DA_PCC_30ndoes}
    \end{subfigure}
    \caption{The PCC plots for CN with ENKF and ETKF in inference. The x-axis stands for different noise levels for model and observation, and the y-axis is the PCC value. Suffix stands for the observation frequency. For example, ENKF\_100 means the model couples the ENKF, and observes the half states with a specific noise level every 100-time steps.}\label{Fig:DA_PCC}
 \noindent\rule{1\linewidth}{1pt}\\
\end{figure*}

\begin{figure*}[t!]
    \centering
    \begin{subfigure}[t]{0.45\textwidth}
        \centering
        \includegraphics[width= 1\linewidth]{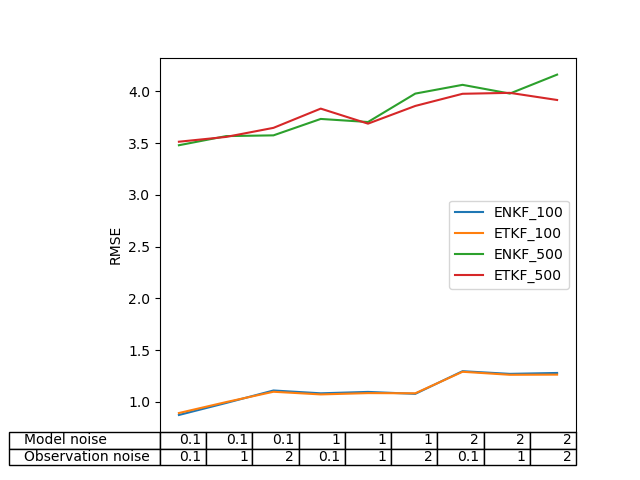} 
        \caption{RMSE for 10 floes situation with ENKF and ETKF}\label{Fig:DA_RMSE_10ndoes}
    \end{subfigure}%
    ~ 
    \begin{subfigure}[t]{0.45\textwidth}
        \centering
        \includegraphics[width= 1\linewidth]{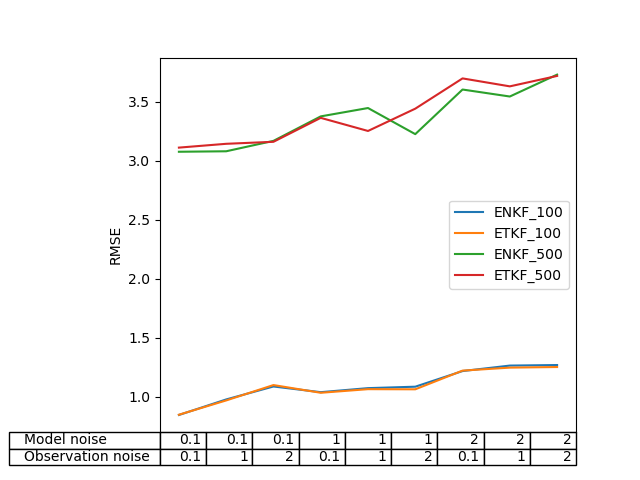}
        \caption{RMSE for 30 floes situation with ENKF and ETKF}\label{Fig:DA_RMSE_30ndoes}
    \end{subfigure}
    \caption{The RMSE plots for CN with ENKF and ETKF in inference. The x-axis stands for different noise levels for model and observation, and the y-axis is the RMSE value. Suffix stands for the observation frequency. For example, ENKF\_100 means the model couples the ENKF, and observes the half states with a specific noise level every 100-time steps.}\label{Fig:DA_RMSE}
 \noindent\rule{1\linewidth}{1pt}\\
\end{figure*}

\subsection*{Data assimilation}

To further enhance the accuracy of this model, the next step involves integrating data assimilation techniques, which will leverage observational data to refine the predictions and improve overall model performance. Incorporating data assimilation into the proposed model is crucial to address the inherent challenge of error accumulation during simulations. Even minor inaccuracies in model states can magnify as simulations progress, leading to significant deviations from ground truth data. By integrating data assimilation, specifically through Ensemble Kalman filter (EnKF)\cite{ENKF1994, ENKF1998} or Ensemble transform Kalman filter (ETKF)\cite{ETKF2001, ETKF2007}, observational data can systematically be combined with model predictions. This method adjusts the model state by weighting the differences between observed values and model predictions, effectively reducing the forecast error. This approach not only enhances the accuracy and reliability of the model's predictions but also extends its applicative value in practical scenarios, ensuring that it remains robust against the underlying uncertainties inherent in environmental modeling.

The simulation framework developed in this study incorporates data assimilation techniques to address the inevitable accumulation of prediction error in long-term forecasting. The proposed model generates future states in a recurrent manner: only the first two time-step states are required as input for initialization, and for each subsequent step, the model takes the two most recent predicted states to forecast the next. While this design provides flexibility and computational efficiency, it also introduces the potential for error propagation over extended time horizons, as any modeling inaccuracies can compound recursively.

To mitigate this error, the model integrates EnKF and ETKF methodologies into the simulation loop. These data assimilation techniques enable the system to incorporate partial observations (positional data of sea ice floes) at specified intervals to recalibrate the internal model state. which reflects both observational availability and practical constraints in real sea-ice monitoring. In many operational and research settings, remote sensing provides snapshots of ice patterns, and ice motion is inferred by comparing the displacement of identifiable structures between successive images, rather than directly measuring complete dynamical states for every floe \cite{lavergne2021towards, wang2022intercomparison}.

Different model noises and observation noises are injected to represent uncertainty in both the learned simulator and the measurements to better reflect the reality. In the Kalman filtering formulation, the model is assumed imperfect with additive process noise, and observations are assumed noisy with additive observation noise; the analysis step then combines model forecasts and measurements using weights determined by their relative uncertainties. In this probabilistic framework, a larger specified model error increases the degree to which the analysis corrects toward observations, whereas a larger observation error yields analyses that remain closer to the model forecast.

In the data assimilation approach for sea ice dynamics, the choice to use position data as the observational input for the ENKF and ETKF is primarily driven by a decision to align more closely with real-world conditions. The harsh and remote nature of polar environments imposes significant logistical challenges, which often restrict the types of data that can be consistently and reliably gathered. Position data are predominantly accessible through satellite imagery and aerial photography, which are effective for monitoring vast areas of sea ice \cite{lavergne2021towards, wang2022intercomparison}. Thus, the observation model is a linear model, and it observes only the position information and cannot observe any velocity information. For the total 10000-time steps, the observation occurs every 100-time steps, or 500-time steps. When there is no observation, each ensemble member only applies the prediction model. Also, the well-trained proposed model is used as the prediction model for ENKF and ETKF. This ensures that the simulations are both scientifically sound and practically relevant, adhering closely to the realities of polar research and observation capabilities.

The results of data assimilation experiments using both the ENKF and the ETKF within the proposed model clearly demonstrate the critical impact of observation frequency on model performance. As illustrated in the results, a decrease in observation frequency from every 100 to 500-time steps significantly compromises the performance, evidenced by a dramatic increase in error and a sharp decline in the PCC for scenarios involving both 10 and 30 floes (Fig.\ref{Fig:DA_PCC} and Fig.\ref{Fig:DA_RMSE}). Further, even when some level of noise is introduced into the model and observation data, adjusting the observation frequency appropriately allows the model to manage these inaccuracies and still achieve high levels of accuracy (Fig.\ref{Fig:DA_PCC} and Fig.\ref{Fig:DA_RMSE}). This capability is particularly evident in more complex scenarios involving 30 floes, where the appropriate observation frequency setting enables the model to effectively maintain reliability and accuracy against noises. The underlying strength of the data assimilation process provides a robust framework that enhances the model's ability to predict in long time trajectory against noises.

\section{Conclusions}

This research introduces a novel application of proposed CN combined with DA techniques to improve the simulation of sea ice dynamics, mainly focusing on collision processes in one-dimensional settings. The findings indicate that the GNN-based model is capable of accurately simulating the dynamics of sea ice, adhering to physical laws, and handling data from various sources, thus confirming the robustness and versatility of the proposed approach. Moreover, the model significantly reduces computational demands compared to traditional numerical methods, facilitating faster simulations without sacrificing accuracy. However, a notable limitation of the current implementation is its restriction to one-dimensional simulations. This constraint simplifies the modeling but does not capture the full complexity of sea ice dynamics, which naturally occur in a two-dimensional space. The simplification to one dimension might overlook certain interactions and behaviors pertinent to the realistic depiction of sea ice processes. 

Extending the present one-dimensional framework to two dimensions introduces rotational degrees of freedom and frictional contact physics that fundamentally change what a learned simulator must represent: in two dimensional settings, floes can rotate, and collisions generate not only normal impulses but also tangential contact forces that drive sliding–sticking behavior and produce torques, which cannot be captured if the model predicts translation alone. This structure appears explicitly in established floe-scale DEM formulations, where translational momentum includes both contact-normal and contact-tangential forces, and angular momentum includes contact torques proportional to the lever arm crossed with the tangential contact force. Relative velocity and constraint forces have both normal and tangential components and are coupled to angular velocities and torques, making rotation and friction inseparable from collision modeling \cite{damsgaard2021effects, tuhkuri2018review, ApplicationDEM}. Consequently, a GNN-based model for two-dimensional floe collisions should model rotation and translation at the same time by augmenting node states with orientation and angular velocity and by learning edge-to-node torque transfer in addition to force transfer. 

Given the promising results achieved with the one-dimensional model, a logical extension of this research would be to adapt and test the GNN-based model within a two-dimensional framework. This expansion would allow the model to handle more complex interactions and provide a more comprehensive understanding of sea ice dynamics. A two-dimensional model would be able to simulate a broader range of physical phenomena and could potentially offer more precise predictions in more realistic situations.
In summary, while the current model represents a fundamental step toward applying GNNs to the simulation of sea ice dynamics, its extension to two dimensions represents a crucial next step in developing a comprehensive tool that can further effectively contribute to the field of climate research.

\section*{Declarations}
\subsection*{Data availability}

All codes used to produce the raw data are available at the following repository: \href{ https://doi.org/10.6084/m9.figshare.28658180}{figshare}\cite{ZHU:data:2025}. The details of implementation can be found in ``Methods" and ``Supplementary materials".

\subsection*{Code availability}

All codes used to produce the analysis are available at the following repository: \href{ https://doi.org/10.6084/m9.figshare.28658180}{figshare}\cite{ZHU:data:2025}. The details of implementation can be found in ``Methods" and ``Supplementary materials".

\subsection*{Author contributions statement}

R.Z. contributed to all aspects of the article, including the conception of the research idea, the development and design of the study, data analysis, and the drafting and revision of the manuscript.

\subsection*{Correspondence} Correspondence and requests for materials should be addressed to Ruibiao Zhu.

\subsection*{Acknowledgements}

R. Zhu gratefully acknowledges the financial support from the Australian Government Research Training Program (RTP) scholarship, as well as the computation support from project zv32 in National Computational Infrastructure (NCI).

\bibliography{OneD}

\begin{appendices}
\section{Supplementary information: kalman filter (KF)}
\label{appendix:KF}

The observation operator $H$ is intentionally partial in both Ensemble kalman filter (EnKF) and Ensemble transform Kalman filter (ETKF). First, only half of the state variables were observed at each assimilation time. This choice reflects a standard geophysical data-assimilation setting where the observation dimension is substantially smaller than the full system state, i.e., only a subset of variables and locations are measured, and the filter must infer the remainder through the forecast–analysis coupling. Second, we assimilated positions only (floe locations), while treating velocities as unobserved. This design is motivated by observational availability and practical constraints in sea-ice monitoring: satellite remote sensing provides widespread spatial coverage but does so primarily through sequential imagery, where motion (and hence velocity) is typically derived by tracking ice features between images rather than directly measured as a state variable \cite{lavergne2021towards}. Consequently, using positions as the primary assimilated quantity better matches the information content and reliability of common observational pipelines, while allowing the filter to infer velocity implicitly through the dynamical model and ensemble cross-covariances. This position-only, partially observed setup therefore emphasizes practical realism and provides a more faithful test of whether the proposed model remains effective when the observation stream is sparse, noisy, and incomplete—conditions that are typical in operational polar data assimilation \cite{wang2022intercomparison}.

All data assimilation experiments are conducted using an ensemble size of $I=100$ for both EnKF and ETKF, with identical settings to ensure a controlled comparison. Covariance inflation is 1 for ETKF. In the experiments, model and observation error covariance matrices $Q$ and $R$ are constructed as diagonal (square) covariance matrices: all off-diagonal entries are set to zero, and diagonal entries are set to $\sigma^2$ with $\sigma\in \{0.1,1,2\}$, corresponding to low-, moderate-, and high-noise regimes. For each ensemble member $i\in I$, the model perturbation $q_i$ and observation perturbation $r_i$ are independently sampled from multivariate normal distributions $q_i\sim\mathcal{N}(0,Q)$ and $r_i\sim\mathcal{N}(0,R)$ with $0$ mean.

\subsection{Ensemble kalman filter (EnKF)}
The Ensemble Kalman Filter is a Monte Carlo method to handle the state estimation problem in high-dimensional systems where the traditional KF becomes computationally impractical \cite{ENKF1994, ENKF1998}. It uses a collection (ensemble) of system states to represent the probability distribution of the state estimates. Each member of the ensemble is updated based on the measurement and the mean and covariance of the ensemble. Compared to Extended Kalman Filter (EKF) \cite{liu2025enhanced}, EnKF does not need any tangent approximation of the nonlinear prediction function. This approach allows for better handling of the uncertainties and nonlinearity in the system by using multiple parallel instances of the filter \cite{ENKF1994, ENKF1998}. 
The Ensemble Kalman filter (EnKF) works as follows: \\

\begin{algorithm}
\caption{Ensemble Kalman filter}\label{alg:EnKF}
\begin{algorithmic}
\Require \text{Known prediction model} $g$ \\
\text{and known observation model} $H$
\State $Q,R,T,I,n \gets \text{Initialized}$
\State $t_j = j\frac{T}{n}, j \in [1, \dots,n]$
    \While{$i \leq I$}
    \State $\bm{u}^i \gets \text{Initialized}$ 
    \EndWhile

\While{$t_j \leq T$}
\If{There is an observation}
    \State $i \gets 1$
    \While{$i \leq I$}
    \State $\bm{q}_i \sim Q$ 
    \State $\bm{u}_{t_{j}}^i \gets  \bm{u}_{t_{j}}^i + \bm{q}_i$
    \EndWhile
    \State $\bm{\bar {u}}_{t_{j}} \gets \frac{1}{I}\sum_{i=1}^{I}\bm{u}_{t_{j}}^i$

\State $U \gets	\begin{bmatrix}
\bm{u}_{t_{j}}^1 - \bm{\bar {u}}_{t_{j}}, & \bm{u}_{t_{j}}^2 - \bm{\bar {u}}_{t_{j}}, & ... & \bm{u}_{t_{j}}^I - \bm{\bar {u}}_{t_{j}}\\
\end{bmatrix} $

\If{The observation model $H$ is nonlinear}
\State \resizebox{0.8\columnwidth}{!}{$V \gets 	\begin{bmatrix}
H(\bm{u}_{t_{j}}^1) - H(\bm{\bar {u}}_{t_{j}}), & ... & H(\bm{u}_{t_{j}}^I) - H(\bm{\bar {u}}_{t_{j}})\\
\end{bmatrix}$}
\ElsIf{The observation model $H$ is linear}
    \State $V \gets HU$
    \EndIf

\State $K \gets \frac{1}{I-1}UV^T(\frac{1}{I-1}VV^T+R)^{-1} $
    \State $\bm{y}_{t_j} \gets \text{observation data}$
    \State $i \gets 1$
    \While{$i \leq I$}
    \State $\bm{r}_i \sim R$ 
    \State $\bm{y}_{t_j}^i \gets \bm{y}_{t_j} + \bm{r}_i $
    \State $\bm{u}_{t_j}^i \gets \bm{u}_{t_j}^i + K(\bm{y}_{t_j}^i - H(\bm{u}_{t_j}^i))$ 
    
    \EndWhile
\ElsIf{There is no observation}
    \State $i \gets 1$
    \While{$i \leq I$}
    \State $\bm{q}_i \sim Q$ 
    \State $\bm{u}_{t_{j}}^i \gets g(\bm{u}_{t_{j}}^i) + \bm{q}_i$ 
    \EndWhile
\EndIf
\EndWhile
\State $\bm{\bar u} \gets \frac{1}{I}\sum_{i=1}^{I}\bm{u}^i$ 
\end{algorithmic}
\end{algorithm}
\subsubsection{Prediction step}
Let $\bm{u}_{t_{j}}$ be one of the ensemble members at time $t_{j}$. If there are $I$ ensemble members, $\bm{u}_{t_{j}}^i =  \bm{u}_{t_{j}}^i + \bm{q}_i$ where $\bm{q}_i$ samples from Q, and the ensemble mean $\bm{\bar {u}}_{t_{j}}$ would be $\bm{\bar {u}}_{t_{j}} = \frac{1}{I}\sum_{i=1}^{I}\bm{u}_{t_{j}}^i$.

The matrix $U \in R^{N \times I}$ and $V \in R^{M \times I}$ are formed, where $N$ is the number of states in the prediction model, and $M$ is the number of states in the observation model,
$$U = 	\begin{bmatrix}
\bm{u}_{t_{j}}^1 - \bm{\bar {u}}_{t_{j}}, & \bm{u}_{t_{j}}^2 - \bm{\bar {u}}_{t_{j}}, & ... & \bm{u}_{t_{j}}^I - \bm{\bar {u}}_{t_{j}}\\
\end{bmatrix},$$ and \\
\resizebox{\columnwidth}{!}{
$V = 	\begin{bmatrix}
H(\bm{u}_{t_{j}}^1) - H(\bm{\bar {u}}_{t_{j}}), & H(\bm{u}_{t_{j}}^2) - H(\bm{\bar {u}}_{t_{j}}), & ... & H(\bm{u}_{t_{j}}^I) - H(\bm{\bar {u}}_{t_{j}})\\
\end{bmatrix},$} \\where $H$ is the nonlinear observation operator. 
When observation operator is linear called $H$, $$V = HU$$
\subsubsection{Update step}
Compute the Kalman Gain
\begin{equation}
    K = \frac{1}{I-1}UV^T(\frac{1}{I-1}VV^T+R)^{-1} 
\end{equation}

Perturb the observation to approximate the posterior error covariance matrix \cite{ENKF1998},
\begin{equation}
    \bm{y}_{t_j}^i = \bm{y}_{t_j} + \bm{r}_i
\end{equation}  where $\bm{y}_{t_j}$ is the observation at time $t_j$ and $\bm{r}_i$ samples from $R$.
Update the estimate via measurement
\begin{equation}
     \bm{u}_{t_j}^i =  \bm{u}_{t_j}^i + K(\bm{y}_{t_j}^i - H(\bm{u}_{t_j}^i))
\end{equation}

\subsection{Ensemble transform kalman filter (ETKF)}
The Ensemble Transform Kalman Filter is a variant of the ensemble Kalman filter \cite{ETKF2001, ETKF2007}. It focuses on transforming the ensemble members to better align with the measurement update. This transformation ensures that the ensemble maintains the correct statistical properties (mean and covariance) after the update step. The ETKF uses a deterministic approach to generate the ensemble members, making it more efficient in specific applications, particularly those involving geophysical and meteorological forecasting \cite{ETKF2001, ETKF2007}.

The Ensemble transform Kalman filter (ETKF) works as follows: \\

\begin{algorithm}
\caption{Ensemble transform Kalman filter}\label{alg:ETKF}
\begin{algorithmic}
\Require \text{Known prediction model} $g$ \\
\text{and known observation model} $H$
\State $Q,R,T,I,n \gets \text{Initialised}$
\State $t_j \gets j\frac{T}{n}, j \in [1, \dots,n]$
    \While{$i \leq I$}
    \State $\bm{u}^i \gets \text{Initialized}$ 
    \EndWhile
\While{$t_j  \leq T$}
\If{There is an observation}
    \State $i \gets 1$
    \While{$i \leq I$}
    \State $\bm{q}_i \sim Q$ 
    \State $\bm{u}_{t_{j}}^i \gets  \bm{u}_{t_{j}}^i + \bm{q}_i$
    \EndWhile

\State $\bm{\bar {u}}_{t_{j}} \gets \frac{1}{I}\sum_{i=1}^{I}\bm{u}_{t_{j}}^i$

\State $U \gets	\begin{bmatrix}
\bm{u}_{t_{j}}^1 - \bm{\bar {u}}_{t_{j}}, & \bm{u}_{t_{j}}^2 - \bm{\bar {u}}_{t_{j}}, & ... & \bm{u}_{t_{j}}^I - \bm{\bar {u}}_{t_{j}}\\
\end{bmatrix} $

\If{The observation model $H$ is nonlinear}
\State \resizebox{0.8\columnwidth}{!}{$V \gets 	\begin{bmatrix}
H(\bm{u}_{t_{j}}^1) - H(\bm{\bar {u}}_{t_{j}}),  & ... & H(\bm{u}_{t_{j}}^I) - H(\bm{\bar {u}}_{t_{j}})\\
\end{bmatrix}$}
\ElsIf{The observation model $H$ is linear}
    \State $V \gets HU$
    \EndIf

    \State $J \gets \frac{I-1}{1 + r}I + V^T(R)^{-1}V$ 
    \State $X\Sigma X^T \gets \text{svd}(J)$\Comment{Singular value decomposition } 
    \State $K \gets U(J)^{-1}V^T(R)^{-1}  $ 
    \State $T \gets \sqrt {I-1} X\Sigma^{-\frac{1}{2}} X^T$\Comment{The transformation matrix}
    \State $\bm{y}_{t_j} \gets \text{observation data}$
    \State $\bm{\bar {u}}_{t_{j}} \gets \bm{\bar {u}}_{t_{j}} + K(\bm{y}_{t_j} - H(\bm{\bar {u}}_{t_{j}}))$
    \State $U \gets UT$
    \State $i \gets 1$
    \While{$i \leq I$}
    \State $\bm{u}_{t_{j}}^i \gets \bm{u}_{t_{j}}^i + \bm{\bar {u}}_{t_{j}} $ 
    \EndWhile

\ElsIf{There is no observation}
    \State $i \gets 1$
    \While{$i \leq I$}
    \State $\bm{q}_i \sim Q$ 
    \State $\bm{u}_{t_{j}}^i \gets g(\bm{u}_{t_{j}}^i) + \bm{q}_i$ 
    \EndWhile
\EndIf
\EndWhile
\State $\bm{\bar u} \gets \frac{1}{I}\sum_{i=1}^{I}\bm{u^i}$ 
\end{algorithmic}
\end{algorithm}

\subsubsection{Prediction step}
Let $\bm{u}_{t_{j}}$ be one of the ensemble members at time $t_{j}$. If there are $I$ ensemble members, $\bm{u}_{t_{j}}^i =  \bm{u}_{t_{j}}^i + \bm{q}_i$ where $\bm{q}_i$ samples from Q, and the ensemble mean $\bm{\bar {u}}_{t_{j}}$ would be $\bm{\bar {u}}_{t_{j}} = \frac{1}{I}\sum_{i=1}^{I}\bm{u}_{t_{j}}^i$.

The matrix $U \in R^{N \times I}$ and $V \in R^{M \times I}$ are formed, where $N$ is the number of states in the prediction model, and $M$ is the number of states in the observation model,
$$U = 	\begin{bmatrix}
\bm{u}_{t_{j}}^1 - \bm{\bar {u}}_{t_{j}}, & \bm{u}_{t_{j}}^2 - \bm{\bar {u}}_{t_{j}}, & ... & \bm{u}_{t_{j}}^I - \bm{\bar {u}}_{t_{j}}\\
\end{bmatrix},$$ and \\
\resizebox{\columnwidth}{!}{
$V = 	\begin{bmatrix}
H(\bm{u}_{t_{j}}^1) - H(\bm{\bar {u}}_{t_{j}}), & H(\bm{u}_{t_{j}}^2) - H(\bm{\bar {u}}_{t_{j}}), & ... & H(\bm{u}_{t_{j}}^I) - H(\bm{\bar {u}}_{t_{j}})\\
\end{bmatrix},$} \\ where $H$ is the nonlinear observation operator. 
When observation operator is linear called $H$, $$V = HU$$

\subsubsection{Update step}

Compute the singular value decomposition of the $I \times I$ matrix
\begin{equation} J = \frac{I-1}{1 + r}I + V^T(R)^{-1}V = X\Sigma X^T\end{equation}
where $r$ is the inflation number.

Compute the Kalman Gain
\begin{equation}
     K = U(X\Sigma^{-1} X^T)V^T(R)^{-1} 
\end{equation}

Update the ensemble mean 
\begin{equation}  \bm{\bar {u}}_{t_j} = \bm{\bar {u}}_{t_j} + K(\bm{y}_{t_j} - H(\bm{\bar {u}}_{t_j}))\end{equation}

Compute the transformation matrix
\begin{equation} T = \sqrt {I-1} X\Sigma^{-\frac{1}{2}} X^T\end{equation}

Update posterior ensemble
\begin{equation}
     U = UT
\end{equation}
\begin{equation}
     \bm{u}_{t_j}^i =  \bm{u}_{t_j}^i + \bm{\bar {u}}_{t_j}
\end{equation}

\subsection{Results for CN with DA in 10 floes and 30 floes situation}
\label{appendix:DA_results}
The following tables are the PCC and RMSE for ENKF and ETKF in 10 floes and 30 floes situations with different noise levels. Suffix 100 and 500 means the proposed model observes the half states every 100 and 500 time steps, respectively.

\begin{table*}[ht]
\caption{PCC for ENKF and ETKF in 10 floes with different noise levels}
\label{table:pcc_10nodes}
\resizebox{\linewidth}{!}{
\begin{tabular*}{\textwidth}{@{\extracolsep\fill}cccccc} \toprule
& & \multicolumn{4}{@{}c@{}}{PCC}\\
\cmidrule{3-6}
    {Model noise} & {Observation noise} & {ENKF 100} & {ETKF 100} & {ENKF 500}  & {ETKF 500} \\ \midrule
0.1 & 0.1 & 0.99348 & 0.99313 & 0.90178 & 0.89945\\
0.1 & 1.0 & 0.99220 & 0.99207 & 0.89765 & 0.89774\\
0.1 & 2.0 & 0.99042 & 0.99073 & 0.89908 & 0.89312\\ \midrule
1.0 & 0.1 & 0.99040 & 0.99057 & 0.88816 & 0.88732\\
1.0 & 1.0 & 0.99005 & 0.99017 & 0.89026 & 0.89120\\
1.0 & 2.0 & 0.99042 & 0.99048 & 0.87418 & 0.88799\\ \midrule
2.0 & 0.1 & 0.98672 & 0.98621 & 0.86818 & 0.87926\\
2.0 & 1.0 & 0.98722 & 0.98700 & 0.88694 & 0.88400\\
2.0 & 2.0 & 0.98696 & 0.98695 & 0.87385 & 0.88544\\ \bottomrule
\end{tabular*}
}
\end{table*}

\begin{table*}[ht]
\caption{RMSE for ENKF and ETKF in 10 floes with different noise levels}
\label{table:rmse_10nodes}
\resizebox{\linewidth}{!}{
\begin{tabular*}{\textwidth}{@{\extracolsep\fill}cccccc} \toprule
& & \multicolumn{4}{@{}c@{}}{RMSE}\\
\cmidrule{3-6}
{Model noise} & {Observation noise} & {ENKF 100}  & {ETKF 100} & {ENKF 500} & {ETKF 500} \\ \midrule
0.1 & 0.1 & 0.87184 & 0.89075 & 3.47952 & 3.51363\\
0.1 & 1.0 & 0.98809 & 0.99665 & 3.56790 & 3.56005\\
0.1 & 2.0 & 1.10862 & 1.09638 & 3.57505 & 3.64760\\\midrule
1.0 & 0.1 & 1.08139 & 1.07064 & 3.73382 & 3.83360\\
1.0 & 1.0 & 1.09488 & 1.08325 & 3.70420 & 3.68827\\
1.0 & 2.0 & 1.07619 & 1.08163 & 3.97873 & 3.85944\\\midrule
2.0 & 0.1 & 1.29509 & 1.28958 & 4.06347 & 3.97697\\
2.0 & 1.0 & 1.26932 & 1.26045 & 3.97917 & 3.98605\\
2.0 & 2.0 & 1.27806 & 1.26198 & 4.16228 & 3.91707\\ \bottomrule

\end{tabular*}}
\end{table*}

\begin{table*}[ht]
\caption{PCC for ENKF and ETKF in 30 floes with different noise levels}
\label{table:pcc_30nodes}
\resizebox{\linewidth}{!}{
\begin{tabular*}{\textwidth}{@{\extracolsep\fill}cccccc} \toprule 
& & \multicolumn{4}{@{}c@{}}{PCC}\\
\cmidrule{3-6}
    {Model noise} & {Observation noise} & {ENKF 100} & {ETKF 100} & {ENKF 500}  & {ETKF 500} \\ \midrule
    0.1 & 0.1 & 0.99348 & 0.99135 & 0.89735 & 0.89360\\
0.1 & 1.0 & 0.98932 & 0.98951 & 0.89713 & 0.89262\\
0.1 & 2.0 & 0.98749 & 0.98713 & 0.89187 & 0.89227\\ \midrule
1.0 & 0.1 & 0.98755 & 0.98778 & 0.87743 & 0.88166\\
1.0 & 1.0 & 0.98700 & 0.98739 & 0.87368 & 0.88840\\
1.0 & 2.0 & 0.98686 & 0.98708 & 0.89046 & 0.87620\\ \midrule
2.0 & 0.1 & 0.98361 & 0.98347 & 0.86826 & 0.86095\\
2.0 & 1.0 & 0.98242 & 0.98284 & 0.87036 & 0.86763\\
2.0 & 2.0 & 0.98210 & 0.98239 & 0.86220 & 0.86103\\ \bottomrule
\end{tabular*}}

\end{table*}

\begin{table*}[ht]
\caption{RMSE for ENKF and ETKF in 30 floes with different noise levels}
\label{table:rmse_30nodes}
\resizebox{\linewidth}{!}{
\begin{tabular*}{\textwidth}{@{\extracolsep\fill}cccccc} \toprule
& & \multicolumn{4}{@{}c@{}}{RMSE}\\
\cmidrule{3-6}
    {Model noise} & {Observation noise} & {ENKF 100}  & {ETKF 100} & {ENKF 500} & {ETKF 500} \\ \midrule
0.1 & 0.1 & 0.84700 & 0.84920 & 3.07565 & 3.11070\\
0.1 & 1.0 & 0.97852 & 0.97021 & 3.07969 & 3.14298\\
0.1 & 2.0 & 1.08828 & 1.10031 & 3.16782 & 3.15995\\ \midrule
1.0 & 0.1 & 1.04012 & 1.03504 & 3.37514 & 3.36270\\
1.0 & 1.0 & 1.07427 & 1.06559 & 3.44587 & 3.25165\\
1.0 & 2.0 & 1.08645 & 1.06442 & 3.22479 & 3.44050\\ \midrule
2.0 & 0.1 & 1.21880 & 1.22260 & 3.60253 & 3.69672\\
2.0 & 1.0 & 1.26526 & 1.24798 & 3.54338 & 3.62936\\
2.0 & 2.0 & 1.26987 & 1.25311 & 3.72876 & 3.71761\\\bottomrule
\end{tabular*}}

\end{table*}

\newpage
\section{Supplementary information: sender and receiver relation matrices}
\label{appendix:Relation_Matrices}
\subsection{Definition of sender and receiver relation matrices}

The sender relation matrix $R_s$ and the receiver relation matrix $R_r$ are both binary matrices that represent the connections between nodes in a graph through directed edges. Here is a detailed definition of each, including the indexing of nodes and edges:

\subsubsection{Indexing of nodes and edges}

Before defining the matrices, it is crucial to establish the indexing for nodes and edges:
\begin{itemize}
    \item Nodes in the graph are indexed arbitrarily from $1$ to $n$, where $n$ is the total number of nodes in the graph.
    \item Edges are also indexed arbitrarily from $1$ to $e$, where $e$ is the total number of directed edges in the graph.
    \item The ordering of these indices does not impact the structural or functional properties of the matrices; it only affects the representation.
\end{itemize}

\subsubsection{Sender relation matrix} \label{R_s definition} 

\begin{itemize}
    \item \textbf{Dimensions}: $R_s$ has dimensions $n \times e$, where $n$ is the number of nodes and $e$ is the number of edges in the graph.
    \item \textbf{Entries}: The entries of $R_s$ are binary:
    \begin{itemize}
        \item $R_s[i, j] = 1$ if the $j$-th edge originates from node $i$.
        \item $R_s[i, j] = 0$ if the $j$-th edge does not originate from node $i$.
    \end{itemize}
\end{itemize}

This matrix is used to identify which node is the sender or the starting point for each directed edge in the graph.

\subsubsection{Receiver relation matrix} \label{R_r definition} 

\begin{itemize}
    \item \textbf{Dimensions}: $R_r$ also has dimensions $n \times e$.
    \item \textbf{Entries}: The entries of $R_r$ are binary:
    \begin{itemize}
        \item $R_r[i, j] = 1$ if the $j$-th edge terminates at node $i$.
        \item $R_r[i, j] = 0$ if the $j$-th edge does not terminate at node $i$.
    \end{itemize}
\end{itemize}

This matrix indicates which node is the receiver or the endpoint for each directed edge in the graph.

\subsection{Definition and proof of the column permutation operator in bi-directional graph} 

Given the sender relation matrix \( R_s \) and the receiver relation matrix \( R_r \) for a graph, where each directed edge in the graph is represented by an entry in these matrices, a column permutation operator \( P \) is defined that reorders the columns of one matrix to transform it into the other, under the condition that each relation between two nodes is represented by two directional edges.

\subsubsection{Setup}

\begin{itemize}
    \item \textbf{Graph Definition}: Assume a graph with nodes and directed edges such that each relationship between two nodes \( u \) and \( v \) is represented by two directed edges: one from \( u \) to \( v \) and the other from \( v \) to \( u \).
    \item \textbf{Matrix Dimensions}: \( R_s \) and \( R_r \) are \( n \times e \) matrices, where \( n \) is the number of nodes and \( e \) is the number of directed edges, $e$ is not necessarily equals to $(n-1)\times2$ since there are some nodes might not be connected \cite{barabasi2013network}.
\end{itemize}

\subsubsection{Column permutation operator} 

\begin{itemize}
    \item \textbf{Definition}: Column permutation operator \( P \) is an \( e \times e \) permutation matrix, where each row and each column contains exactly one entry of 1, and all other entries are 0.
    \item \textbf{Action}: \( P \) acts on the columns of \( R_s \) such that \( R_sP \) reorders the columns of \( R_s \). The goal is to find a permutation \( P \) such that \( R_sP = R_r \).
\end{itemize}

\subsubsection{Proving the existence of permutation operator} 

To demonstrate that such a permutation matrix \( P \) exists, the prove as follows:

\begin{enumerate}
    \item \textbf{Bijective mapping of edges}: Each edge \( i \) in \( R_s \) representing a directed edge from node \( u \) to node \( v \) has a corresponding edge \( j \) in \( R_r \) representing the directed edge from node \( v \) to node \( u \), given the assumption that each relationship is bidirectional.
    \item \textbf{Construction of \( P \)}:
    \begin{itemize}
        \item For each column \( i \) in \( R_s \) (representing edge \( i: u \to v \)), there exists a unique column \( j \) in \( R_r \) (representing edge \( j: v \to u \)).
        \item Set \( P_{ji} = 1 \) if and only if the edge represented by column \( i \) in \( R_s \) corresponds to the edge represented by column \( j \) in \( R_r \).
    \end{itemize}
    \item \textbf{Validity of \( P \)}:
    \begin{itemize}
        \item Since each directed edge in the graph is unique and bidirectional, the mapping from edges in \( R_s \) to \( R_r \) is bijective (one-to-one and onto). Thus, \( P \) constructed in this manner is a valid permutation matrix.
    \end{itemize}
    \item \textbf{Permutation Result}: Applying \( P \) to \( R_s \), \( R_sP \) is constructed. By the construction of \( P \), \( R_sP \) will have its columns rearranged to match exactly the columns of \( R_r \).
\end{enumerate}
\subsection{Properties of the permutation matrix}
\begin{itemize}
    \item Each row and each column of \( P \) contains exactly one entry of 1.
    \item The matrix \( P \) rearranges (permutes) the columns of any matrix it multiplies.
    \item \( P \) is symmetric, meaning that \( P = P^T \). This property arises because ,for every entry \( P_{ij} = 1 \), where \( i \) is permuted to \( j \), there is a corresponding entry \( P_{ji} = 1 \) reflecting the inverse of that permutation, which, for a permutation matrix that swaps pairs of indices, means \( i \) and \( j \) swap roles symmetrically.
\end{itemize}

\subsection{Multiplication of permutation matrices}
The product \( P \times P \) results in another matrix where each entry \( (i, j) \) is computed by:
\[
(P \times P)_{ij} = \sum_{k=1}^n P_{ik} P_{kj}
\]
This formula considers the intermediate step \( k \), where \( P_{ik} = 1 \) if the permutation moves column \( i \) to column \( k \), and \( P_{kj} = 1 \) if it subsequently moves column \( k \) to column \( j \).

\subsection{Identity matrix} 
The identity matrix \( I \) is defined as:
\[
I_{ij} = 
  \begin{cases} 
    1 & \text{if } i = j \\
    0 & \text{if } i \neq j
  \end{cases}
\]
It leaves any vector or matrix unchanged when multiplied.

\subsection{Proof of relationship between identity matrix and permutation matrix}
By definition, the property has $\sum_{k=1}^n P_{ik} P_{kj} = 0$ if and only if $i = j$. Therefore,
\begin{enumerate}
    \item \textbf{Diagonal elements}: For each \( i \), \( (P \times P)_{ii} = 1 \) because the column \( i \) moves to some column \( k \) and then back to \( i \) due to the permutation, ensuring it returns to its original position.
    \item \textbf{Off-diagonal elements}: For each \( i \neq j \), \( (P \times P)_{ij} = 0 \) because there is no intermediate column \( k \) such that column \( i \) would move to column \( j \) through \( k \). The only contributing \( k \) would have to coincide with both \( i \) and \( j \), which cannot happen if \( i \neq j \).
\end{enumerate}
Thus, \( P \times P = I \).

\begin{figure*} 
\centering
\subfloat[A graph]{\includegraphics[width=0.8\columnwidth]{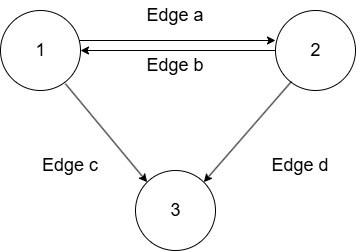}\label{Fig:general_edge_nodes}}
\hfil
\subfloat[A bi-directional graph]{\includegraphics[width=0.8\columnwidth]{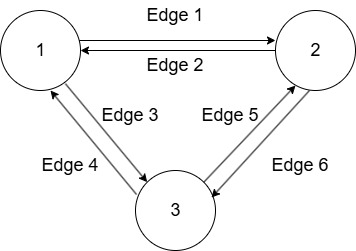}\label{Fig:edge_nodes}}
\caption{Illustration of constructing  graphs (Circle represents node, and arrow represents edge)}
\label{Fig:general_edge_nodes2}
\end{figure*}

\subsubsection{Theoretical result}

Thus, the column permutation matrix \( P \) exists such that \( R_sP = R_r \), under the assumption of bidirectional relationships represented as pairs of directed edges. This shows that \( P \) correctly translates the sender matrix into the receiver matrix by simply reordering the edges according to their reverse directions. 

\begin{lemma}
Given relation matrices $R_s$ and $R_r$ for a graph where a relation between two nodes is represented as two directional edges, there exists a column permutation of one of the matrices to represent the other one. The permutation would be column $i$ and column $j$ when two directional edges are indexed as  $i$ and $j$ between two nodes.
\end{lemma}

Utilizing column permutation to represent $R_s$ is particularly advantageous in terms of memory storage. Instead of storing a completely new matrix $R_s$, the matrix $R_r$ can be simply stored, and $R_s$ matrix can be derived by column permutation. 

For example, if the graph is constructed as Figure \ref{Fig:edge_nodes}, the relation matrices $R_s$ and $R_r$ would be as follows:

\[\resizebox{\hsize}{!}{
$R_s  = 
\begin{blockarray}{cccccccc}
 & \text{edge}_1 & \text{edge}_2 & \text{edge}_3 & \text{edge}_4 & \text{edge}_5 & \text{edge}_6 \\
\begin{block}{c[cccccc]c}
\text{node}_1 & 1 & 0 & 1 & 0 & 0 & 0 \bigstrut[t] & =2 \\
\text{node}_2 & 0 & 1 & 0 & 0 & 0 & 1\bigstrut[t] & =2 \\
\text{node}_3 & 0 & 0 & 0 & 1 & 1 & 0\bigstrut[t] & =2 \\
\end{block}
 & = 1 & = 1 & = 1 & = 1 & = 1 & = 1  \\
\end{blockarray}$}
\]

\[\resizebox{\hsize}{!}{$
R_r  = 
\begin{blockarray}{cccccccc}
 & \text{edge}_1 & \text{edge}_2 & \text{edge}_3 & \text{edge}_4 & \text{edge}_5 & \text{edge}_6 \\
\begin{block}{c[cccccc]c}
\text{node}_1 & 0 & 1 & 0 & 1 & 0 & 0 \bigstrut[t] & =2 \\
\text{node}_2 & 1 & 0 & 0 & 0 & 1 & 0\bigstrut[t] & =2 \\
\text{node}_3 & 0 & 0 & 1 & 0 & 0 & 1\bigstrut[t] & =2 \\
\end{block}
 & = 1 & = 1 & = 1 & = 1 & = 1 & = 1  \\
\end{blockarray}$}
\]

It can be identified that the columns 2, 4 and 6 permutations of $R_r$ into columns 1, 3 and 5 would be identical with $R_s$. Additionally, the column permutation matrix can be obtained by definition as follows:
\[ P =
     \begin{bmatrix}
         0 & 1 & 0 & 0 & 0 & 0\\
         1 & 0 & 0 & 0 & 0 & 0\\
         0 & 0 & 0 & 1 & 0 & 0\\
         0 & 0 & 1 & 0 & 0 & 0\\
         0 & 0 & 0 & 0 & 0 & 1\\
         0 & 0 & 0 & 0 & 1 & 0
     \end{bmatrix} \]

The result can be verified by checking \[ R_s \times P = R_r, \] as follows: 
\[\resizebox{\hsize}{!}{$
\begin{bmatrix}
          1 & 0 & 1 & 0 & 0 & 0\\
         0 & 1 & 0 & 0 & 0 & 1\\ 
         0 & 0 & 0 & 1 & 1 & 0
     \end{bmatrix}
     \times
     \begin{bmatrix}
         0 & 1 & 0 & 0 & 0 & 0\\
         1 & 0 & 0 & 0 & 0 & 0\\
         0 & 0 & 0 & 1 & 0 & 0\\
         0 & 0 & 1 & 0 & 0 & 0\\
         0 & 0 & 0 & 0 & 0 & 1\\
         0 & 0 & 0 & 0 & 1 & 0
     \end{bmatrix}
      = 
     \begin{bmatrix}
         0 & 1 & 0 & 1 & 0 & 0\\
         1 & 0 & 0 & 0 & 1 & 0\\ 
         0 & 0 & 1 & 0 & 0 & 1
     \end{bmatrix} $}\]
\begin{remark}
Since $P$ is an involutory matrix, both $$R_s \times P = R_r$$ and 
$$R_r \times P = R_s $$ holds. Also, $P_{ij} = 1$ if and only if labeling a pair of edges as $i$ and $j$.
\end{remark}

Thus, applying the permutation requires far less memory than storing an entirely new matrix.  

For the 10 floes setup as Fig.\ref{fig:floe_edge_setup},   the same definitions are used from \ref{R_s definition} and \ref{R_r definition}, and the $R_s$ and $R_r$ can be constructed as the following:

\begin{figure*}
\centering

\[\resizebox{\linewidth}{!}{
$R_s  = 
\begin{blockarray}{ccccccccccccccccccccccc}
 & \text{edge}_1 & \text{edge}_2 & \text{edge}_3 & \text{edge}_4 & \text{edge}_5 & \text{edge}_6
 & \text{edge}_7 & \text{edge}_8 & \text{edge}_9 & \text{edge}_{10} & \text{edge}_{11} & \text{edge}_{12} 
 & \text{edge}_{13} & \text{edge}_{14} & \text{edge}_{15} & \text{edge}_{16} & \text{edge}_{17} & \text{edge}_{18} 
 & \text{edge}_{19} & \text{edge}_{20} & \text{edge}_{21} & \text{edge}_{22} \\
\begin{block}{c[cccccccccccccccccccccc]}
\text{left boundary} & 1 & 0 & 0 & 0 & 0 & 0 & 0 & 0 & 0 & 0 & 0 & 0 & 0 & 0 & 0 & 0 & 0 & 0 & 0 & 0 & 0 & 0\\
\text{x}_1 & 0 & 1 & 1 & 0 & 0 & 0 & 0 & 0 & 0 & 0 & 0 & 0 & 0 & 0 & 0 & 0 & 0 & 0 & 0 & 0 & 0 & 0\\
\text{x}_2 & 0 & 0 & 0 & 1 & 1 & 0 & 0 & 0 & 0 & 0 & 0 & 0 & 0 & 0 & 0 & 0 & 0 & 0 & 0 & 0 & 0 & 0\\
\text{x}_3 & 0 & 0 & 0 & 0 & 0 & 1 & 1 & 0 & 0 & 0 & 0 & 0 & 0 & 0 & 0 & 0 & 0 & 0 & 0 & 0 & 0 & 0\\
\text{x}_4 & 0 & 0 & 0 & 0 & 0 & 0 & 0 & 1 & 1 & 0 & 0 & 0 & 0 & 0 & 0 & 0 & 0 & 0 & 0 & 0 & 0 & 0\\
\text{x}_5 & 0 & 0 & 0 & 0 & 0 & 0 & 0 & 0 & 0 & 1 & 1 & 0 & 0 & 0 & 0 & 0 & 0 & 0 & 0 & 0 & 0 & 0\\
\text{x}_6 & 0 & 0 & 0 & 0 & 0 & 0 & 0 & 0 & 0 & 0 & 0 & 1 & 1 & 0 & 0 & 0 & 0 & 0 & 0 & 0 & 0 & 0\\
\text{x}_7 & 0 & 0 & 0 & 0 & 0 & 0 & 0 & 0 & 0 & 0 & 0 & 0 & 0 & 1 & 1 & 0 & 0 & 0 & 0 & 0 & 0 & 0\\
\text{x}_8 & 0 & 0 & 0 & 0 & 0 & 0 & 0 & 0 & 0 & 0 & 0 & 0 & 0 & 0 & 0 & 1 & 1 & 0 & 0 & 0 & 0 & 0\\
\text{x}_9 & 0 & 0 & 0 & 0 & 0 & 0 & 0 & 0 & 0 & 0 & 0 & 0 & 0 & 0 & 0 & 0 & 0 & 1 & 1 & 0 & 0 & 0\\
\text{x}_{10} & 0 & 0 & 0 & 0 & 0 & 0 & 0 & 0 & 0 & 0 & 0 & 0 & 0 & 0 & 0 & 0 & 0 & 0 & 0 & 1 & 1 & 0\\
\text{right boundary} & 0 & 0 & 0 & 0 & 0 & 0 & 0 & 0 & 0 & 0 & 0 & 0 & 0 & 0 & 0 & 0 & 0 & 0 & 0 & 0 & 0 & 1\\
\end{block}
  \\
\end{blockarray}$}
\]

\[\resizebox{\linewidth}{!}{
$R_r  = 
\begin{blockarray}{ccccccccccccccccccccccc}
 & \text{edge}_1 & \text{edge}_2 & \text{edge}_3 & \text{edge}_4 & \text{edge}_5 & \text{edge}_6
 & \text{edge}_7 & \text{edge}_8 & \text{edge}_9 & \text{edge}_{10} & \text{edge}_{11} & \text{edge}_{12} 
 & \text{edge}_{13} & \text{edge}_{14} & \text{edge}_{15} & \text{edge}_{16} & \text{edge}_{17} & \text{edge}_{18} 
 & \text{edge}_{19} & \text{edge}_{20} & \text{edge}_{21} & \text{edge}_{22} \\
\begin{block}{c[cccccccccccccccccccccc]}
\text{left boundary} &  0 & 1 & 0 & 0 & 0 & 0 & 0 & 0 & 0 & 0 & 0 & 0 & 0 & 0 & 0 & 0 & 0 & 0 & 0 & 0 & 0 & 0\\
\text{x}_1 &  1 & 0 & 0 & 1 & 0 & 0 & 0 & 0 & 0 & 0 & 0 & 0 & 0 & 0 & 0 & 0 & 0 & 0 & 0 & 0 & 0 & 0\\
\text{x}_2 &  0 & 0 & 1 & 0 & 0 & 1 & 0 & 0 & 0 & 0 & 0 & 0 & 0 & 0 & 0 & 0 & 0 & 0 & 0 & 0 & 0 & 0\\
\text{x}_3 &  0 & 0 & 0 & 0 & 1 & 0 & 0 & 1 & 0 & 0 & 0 & 0 & 0 & 0 & 0 & 0 & 0 & 0 & 0 & 0 & 0 & 0\\
\text{x}_4 &  0 & 0 & 0 & 0 & 0 & 0 & 1 & 0 & 0 & 1 & 0 & 0 & 0 & 0 & 0 & 0 & 0 & 0 & 0 & 0 & 0 & 0\\
\text{x}_5 &  0 & 0 & 0 & 0 & 0 & 0 & 0 & 0 & 1 & 0 & 0 & 1 & 0 & 0 & 0 & 0 & 0 & 0 & 0 & 0 & 0 & 0\\
\text{x}_6 &  0 & 0 & 0 & 0 & 0 & 0 & 0 & 0 & 0 & 0 & 1 & 0 & 0 & 1 & 0 & 0 & 0 & 0 & 0 & 0 & 0 & 0\\
\text{x}_7 &  0 & 0 & 0 & 0 & 0 & 0 & 0 & 0 & 0 & 0 & 0 & 0 & 1 & 0 & 0 & 1 & 0 & 0 & 0 & 0 & 0 & 0\\
\text{x}_8 &  0 & 0 & 0 & 0 & 0 & 0 & 0 & 0 & 0 & 0 & 0 & 0 & 0 & 0 & 1 & 0 & 0 & 1 & 0 & 0 & 0 & 0\\
\text{x}_9 &  0 & 0 & 0 & 0 & 0 & 0 & 0 & 0 & 0 & 0 & 0 & 0 & 0 & 0 & 0 & 0 & 1 & 0 & 0 & 1 & 0 & 0\\
\text{x}_{10} &  0 & 0 & 0 & 0 & 0 & 0 & 0 & 0 & 0 & 0 & 0 & 0 & 0 & 0 & 0 & 0 & 0 & 0 & 1 & 0 & 0 & 1\\
\text{right boundary} &  0 & 0 & 0 & 0 & 0 & 0 & 0 & 0 & 0 & 0 & 0 & 0 & 0 & 0 & 0 & 0 & 0 & 0 & 0 & 0 & 1 & 0\\
\end{block}
  \\
\end{blockarray}$}
\]

\end{figure*}

For more realistic two-dimensional case, both $R_s$ and $R_r$ can be constructed with the same matrix definition from \ref{R_s definition} and \ref{R_r definition}. If there are $n$ floes in the two dimensional space, then each floe might interact with at most $n-1$ other floes. Thus, $n(n-1)$ edges are formed and indexed. Both $R_s$ and $R_r$ have $n$ rows and $n(n-1)$ columns, and their values are either 1 or 0, where $1$ represents the edge that exists, and $0$ represents the edge that does not exist. The sum of each row of the matrix is $n-1$, and the sum of each column of the matrix is $1$.     

\section{Supplementary information: Error analysis}
\label{appendix:Error_analysis}
Quantifying the errors in neural networks is an intricate challenge that continues to be an active area of research \cite{cai2023combination}. For example, adding a small noise to perturb the input can result in different classifications in the image \cite{zhang2023generalizing, Goodfellow2014ExplainingAH}. The accuracy of neural network approximations is influenced by multiple factors, including network architecture, the characteristics of the training data, and the specific implementations of learning algorithms. These networks operate within complex, often non-convex optimization landscapes, where the choices of activation functions, network topology, and the training regime all contribute significantly to the network's performance and generalization capabilities outside the training set \cite{GeneralizationFFN2019, NNerror2021}.

Based on the mean value theorem, if $f(x)$ is continuous on $x, y$ and differentiable on $(x, y)$, then $\exists c \in (x, y)$ such that

\begin{equation} 
\begin{split}
f^\prime(c) = \frac{f(y) - f(x)}{y-x}
\end{split}
\end{equation}

If a function $f$ has a continuous derivative $f^\prime$ on $[x, y]$, 
by the mean value theorem, 

\begin{equation} \label{eq:Lipschitz_function}
\begin{split}
||f(y) - f(x)|| \leq C||y-x||
\end{split}
\end{equation}

satisfies a Lipschitz continuous on $[a,b]$ where $C \geq \max|\nabla f(x)|$, which means there exists a constant $C$ that is greater than or equal to the maximum rate of change of function $f$ in its domain. This constant $C$ is called the Lipschitz constant, and this function $f$ is called a Lipschitz function. Thus, the activation function would be chosen as a smooth and differentiable function across the domain to satisfy Lipschitz continuous. The Mish function $Mish(x)$ \cite{misra2019mish} is used for all activation layers in the proposed model for non-monotonic, non-monotonic derivative, not saturated, and infinite continuity. 

\begin{equation} 
\begin{split}
\text{tanh}(x) & = \frac{e^x-e^{-x}}{e^x+e^{-x}} \\
\text{softplus}(x) & = ln(1+e^x) \\
\text{Mish(x)} & = x\text{tanh}(\text{softplus}(x)) \\
& = x\frac{e^{ln(1+e^x)} - e^{-ln(1+e^x)}}{e^{ln(1+e^x)} + e^{-ln(1+e^x)}}
\end{split}
\end{equation}

\begin{equation} \label{eq:mish}
\text{Mish}^{\prime}(x) = \frac{e^x\omega}{\delta^2}
\end{equation}
where $\omega = 4(x+1) + 4e^{2x} + e^{3x} + e^x(4x+6)$ and $\delta = (e^x + 1)^2 +1$.

Based on the chain rule, the composition of smooth functions is also smooth. The proposed model would be smooth after applying all smooth activation functions. 
For predicting the future state, velocity is chosen to be predicted first; then the position is inferred, rather than predicting position directly, based on the following analysis. When directly predicting the node position, an error term of the predicted position is denoted as $\epsilon$. Then, $v_t^i =  \frac{x_t^i + \epsilon - x_{t-1}^i}{dt}$ can be derived. Since the setting $dt = 1e^{-4}$, this would significantly increase the error of predicted position $x_t^i$ to $v_t^i$ with $1e^{4}$ magnitude. Thus, the error would be carried in the next iteration for the velocity feature input as $v_{t+1}^i = v_{t}^i+\epsilon * 1e^4 + \frac{F_{t}^i}{m^i}dt$ for the neural networks, and this causes the predicted position to be inaccurate with more errors. After more iterations, the results would be unreasonable, and the experiments can show the model would simulate unjustifiable results after around a few hundred iterations, which is consistent with the theory. 


The error can be reduced by predicting the velocity rather than the x-axis position in the model since fewer errors in the input of the model can bind tighter output of the model based on the Lipschitz continuous. Let denote $\epsilon_{t}$ is a one-step error for the output of the proposed network with input $\tilde x_{t-1}, \tilde x_{t}, \tilde v_{t}$ at $t$, where $v_{t+1}$ is ground truth velocity value and $\tilde v_{t+1}$ is the predicted velocity value. 

\begin{equation} 
\begin{split}
\epsilon_{t} 
& = \text{CN}(\tilde x_{t-1}, \tilde x_{t}, \tilde v_{t}) - v_{t+1}  \\
& = \tilde v_{t+1} - v_{t+1} 
\end{split}
\end{equation}

Table \ref{table:perdicting comparison} indicates that the model trained to directly predict positions tend to learn a short-horizon mapping from the current state to the next states, rather than learning the underlying physical movement patterns that govern how floe states evolve through time. The CN predicting position even fails to simulate across the whole time domain, thus, the simulation RMSE and PCC are not applicable. Empirically, this distinction becomes most apparent in simulation. While position-predicting variants may achieve good one-step position errors but much worser one-step velocity errors , the errors compound rapidly when the model is run autoregressively, leading to substantially larger simulation error than velocity-predicting counterparts. 
\begin{table*}[ht]
\caption{CN performance comparison with different predicting states}
\label{table:perdicting comparison}
\small
\setlength{\tabcolsep}{4pt} 

\begin{tabular*}{\linewidth}{ccccc} \toprule
{Model} & {Simulation RMSE} & {one-step position RMSE} & {one-step velocity RMSE} & {PCC} \\ 
\midrule
\text{CN predicting velocity} & 2.334024 & 0.0000060 & 0.028410 & 0.934317\\ 
\text{CN predicting position} & - & 0.0000003 & 3.022292 & -\\ 
\bottomrule
\end{tabular*}

\end{table*}

For the predicted position at time $t$, the equation can be expanded into the ground truth position at time $1$, since $\tilde x_2 = x_{1} + \tilde v_{2} dt $ that is predicted by the CN with initial conditions $x_{0}, x_{1}$. Also, $\epsilon_{max}$ is denoted as the maximum one-step prediction error from the proposed model. 

\begin{equation} 
\begin{split}
\tilde x_{t+1} & = \tilde x_{t} + \tilde v_{t+1} dt \\
& = \tilde x_{t} + (v_{t+1} + \epsilon_{t}) dt \\
& = \tilde x_{t-1} + (v_{t} + \epsilon_{t-1}) dt + (v_{t+1} + \epsilon_{t}) dt \\
&  \dots \\
& = \tilde x_2 + (v_3 + \dots + v_{t+1})dt + (\epsilon_{2} + \dots + \epsilon_{t})dt \\
& = x_1 + (v_2 + \dots + v_{t+1})dt + (\epsilon_{1} + \dots +\epsilon_{t})dt \\
& \leq x_1 + (v_2 + \dots + v_{t+1})dt + t\epsilon_{max}dt
\end{split}
\end{equation}

Therefore, the model's error would be linearly bounded by the maximum one-step prediction error, which is preferable to exponentially increasing the error.  

\section{Supplementary information: Skill scores}
\label{appendix:Skill_Scores}
Mean Squared Error (MSE) is a widely used statistical measure for evaluating the performance of predictive models. It has been adopted to evaluate the performance of the model, since MSE is preferred for its ability to heavily penalize larger errors and its suitability for mathematical optimization. 
\begin{equation} \label{eq:MSE}
\text{MSE} = \frac{1}{n}\sum_{i=1}^{n}(\bm{x}_i-\bm{y}_i)^2 
\end{equation}
where $\bm{x}$ is predicted value, $\bm{y}$ is the ground truth value, and $n$ is the size of the sample.

Root Mean Squared Error (RMSE) is computed by taking the square root of MSE.
\begin{equation} \label{eq:RMSE}
    \text{RMSE} = \sqrt{\text{MSE}} = \sqrt{(\frac{1}{n})\sum_{i=1}^{n}(\bm{y}_{i} - \bm{x}_{i})^{2}}
\end{equation}

The pattern correlation coefficient is used to measure the overall strength of the relationship between the predicted trajectories and ground-truth trajectories.
\begin{equation} \label{eq:PCC}
     \text{PCC} =
  \frac{ \sum_{i=1}^{n}(\bm{x}_i-\bar{\bm{x}})(\bm{y}_i-\bar{\bm{y}}) }{%
        \sqrt{\sum_{i=1}^{n}(\bm{x}_i-\bar{\bm{x}})^2}\sqrt{\sum_{i=1}^{n}(\bm{y}_i-\bar{\bm{y}})^2}}
\end{equation}

Different evaluation methods have different attributes and emphasize different traits of the performance. For a more comprehensive and accurate performance analysis, all mentioned techniques will be used to compensate for the limitations of using only a specific method.

\clearpage
\onecolumn
\section{Supplementary information: Training configuration}
\label{appendix:Training}

All models were trained with 10 million ground truth and prediction pairs data with a batch size of 100 using the Adam optimizer \cite{kingma2014adam}, which is widely used due to its computational efficiency and adaptive moment estimates that stabilize stochastic optimization in high-dimensional parameter spaces. The initial learning rate is set to $1\times10^{-4}$ and an exponential learning-rate schedule is applied with a multiplicative decay factor (gamma) of 0.99 per epoch. This choice is motivated by the well-established sensitivity of neural training dynamics to the learning rate. Learning-rate decay is a practice in modern deep learning because it enables larger exploratory updates early in training while progressively shifting toward smaller, fine-grained updates as optimization approaches a solution. Moreover, theoretical and empirical analyses show that maintaining a comparatively larger learning rate early in training, while progressively shifting toward smaller updates, can have a regularization effect and can reduce premature memorization of spurious patterns, thereby improving generalization \cite{d2024we,senior2013empirical,ren2024understanding,li2019towards}.
\begin{figure*}[t!]
    \centering
    \begin{subfigure}[t]{0.45\textwidth}
        \centering
        \includegraphics[width= 1\linewidth]{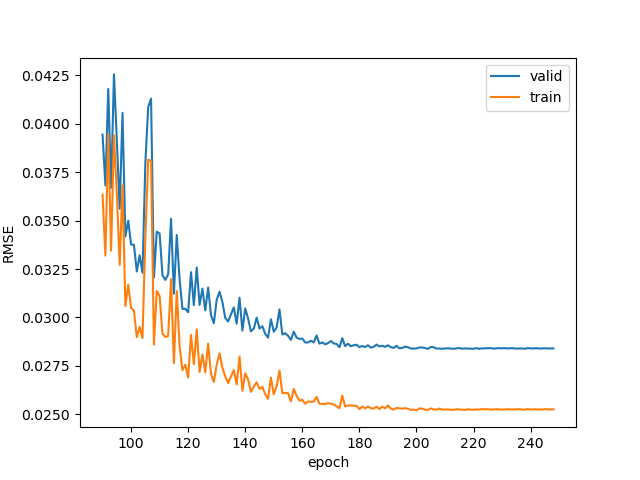} 
        \caption{RMSE for velocity}\label{Fig:velocity_rmse_training}
    \end{subfigure}%
    ~ 
    \begin{subfigure}[t]{0.45\textwidth}
        \centering
        \includegraphics[width= 1\linewidth]{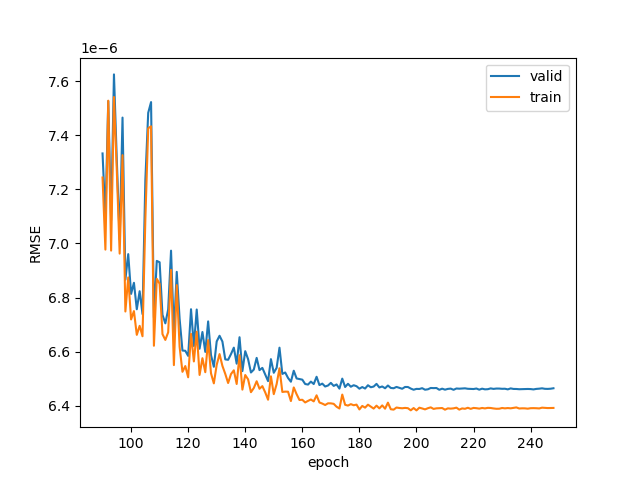}
        \caption{RMSE for position}\label{Fig:position_rmse_training}
    \end{subfigure}
    \caption{Training and validation loss in different epochs.}\label{Fig:training_loss}
 
\end{figure*}

As shown in Figure \ref{Fig:training_loss}, the training and validation losses for both position and velocity stabilize after approximately 180 epochs, with no evidence of divergence between the two curves, and training required approximately 16 hours on a single NVIDIA RTX 4090 GPU. This behavior is consistent with good generalization: overfitting is typically diagnosed when the training loss continues to decrease while the validation loss stops improving and begins to increase, producing a clear separation between training and validation loss trajectories \cite{hawkins2004problem, ying2019overview}. In the experiments, the absence of this characteristic divergence indicates that the selected optimization setting yields stable convergence without observable overfitting under the training duration.

\section{Supplementary information: Visualization}
\label{appendix:Visualization}

To enhance the interpretability of model performance and provide qualitative evidence of physical consistency, supplementary videos are provided with the simulation results. These visualizations are designed to convey critical aspects of the predicted floe dynamics, as the following:

\begin{itemize}
    \item No Unphysical Overlap or Boundary Breach: Throughout the simulations, individual sea ice floes are clearly shown to maintain spatial separation, never unphysical overlapping or passing through one another. This behavior demonstrates that the model correctly maintain physical constraints. Similarly, floes are consistently confined within the simulation domain, and no floe crosses the defined boundary, further validating the adherence to physical domain restrictions.
    \item Realistic Rebound Dynamics: When collisions occur, floes exhibit realistic elastic rebound behavior rather than unphysical sticking or interpenetration. The rebound trajectories are visually consistent with expected outcomes, as defined in the DEM framework.
    
    \item Color-Coded Floe Identification: To support visual tracking, each floe is assigned a distinct color that remains fixed throughout the simulation. This coloring facilitates frame-by-frame analysis of individual floe motion and makes it easier to assess whether floes interact and move correctly.
\end{itemize}

\subsection{Dataset visualization}

\begin{figure}[!htb]
   \begin{minipage}{0.78\textwidth}
     \centering
     \includegraphics[width= 1\textwidth]{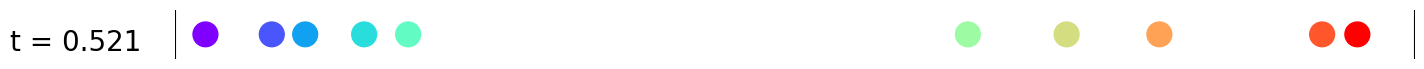}
    \end{minipage}\vfill
\begin{minipage}{0.78\textwidth}
     \centering
     \includegraphics[width= 1\textwidth]{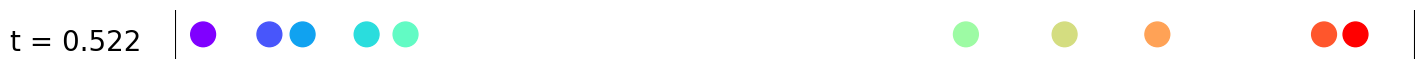}
    \end{minipage}\vfill
    \begin{minipage}{0.78\textwidth}
     \centering
     \includegraphics[width= 1\textwidth]{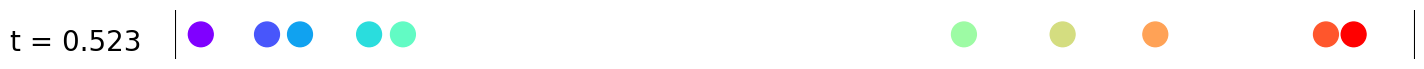}
    \end{minipage}\vfill
    \begin{minipage}{0.78\textwidth}
     \centering
     \includegraphics[width= 1\textwidth]{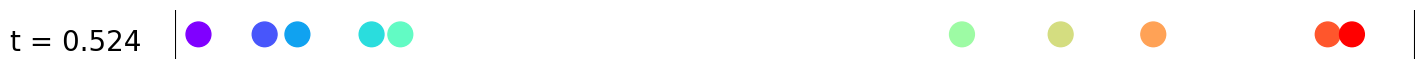}
    \end{minipage}\vfill
    \begin{minipage}{0.78\textwidth}
     \centering
     \includegraphics[width= 1\textwidth]{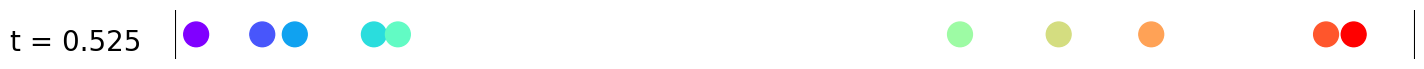}
    \end{minipage}\vfill
    \begin{minipage}{0.78\textwidth}
     \centering
     \includegraphics[width= 1\textwidth]{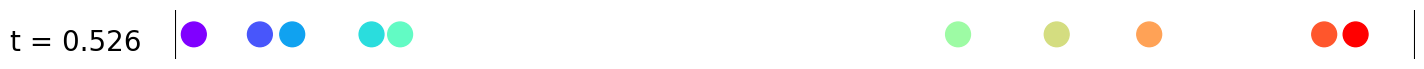}
    \end{minipage}\vfill
    \begin{minipage}{0.78\textwidth}
     \centering
     \includegraphics[width= 1\textwidth]{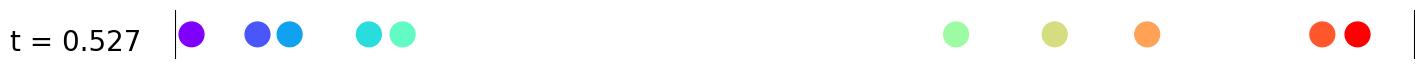}
    \end{minipage}\vfill
    \begin{minipage}{0.78\textwidth}
     \centering
     \includegraphics[width= 1\textwidth]{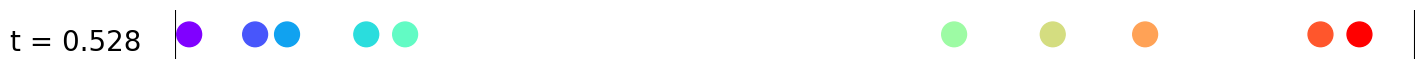}
    \end{minipage}\vfill
    \begin{minipage}{0.78\textwidth}
     \centering
     \includegraphics[width= 1\textwidth]{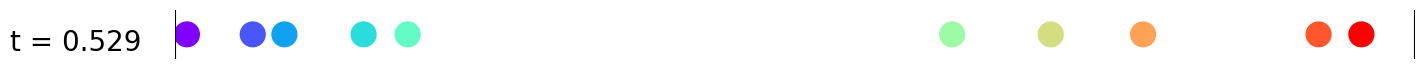}
    \end{minipage}\vfill
    \begin{minipage}{0.78\textwidth}
     \centering
     \includegraphics[width= 1\textwidth]{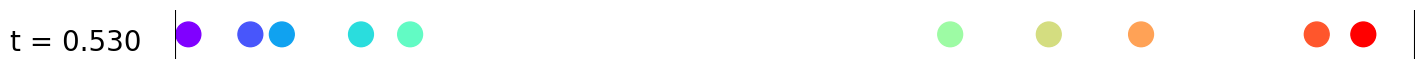}
    \end{minipage}\vfill
    \begin{minipage}{0.78\textwidth}
     \centering
     \includegraphics[width= 1\textwidth]{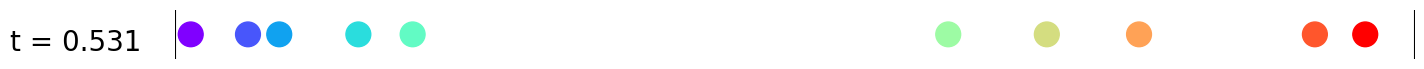}
    \end{minipage}\vfill
\begin{minipage}{0.78\textwidth}
     \centering
     \includegraphics[width= 1\textwidth]{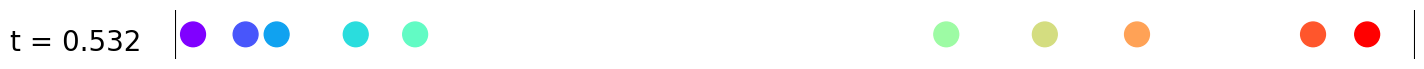}
    \end{minipage}\vfill
    \begin{minipage}{0.78\textwidth}
     \centering
     \includegraphics[width= 1\textwidth]{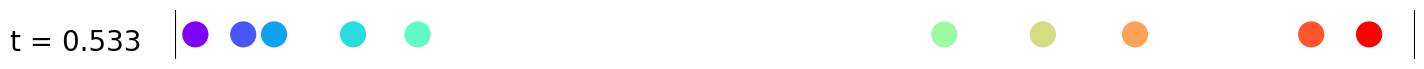}
    \end{minipage}\vfill
    \begin{minipage}{0.78\textwidth}
     \centering
     \includegraphics[width= 1\textwidth]{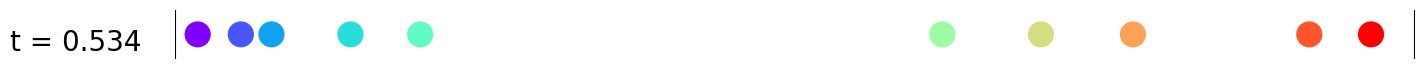}
    \end{minipage}\vfill
    \begin{minipage}{0.78\textwidth}
     \centering
     \includegraphics[width= 1\textwidth]{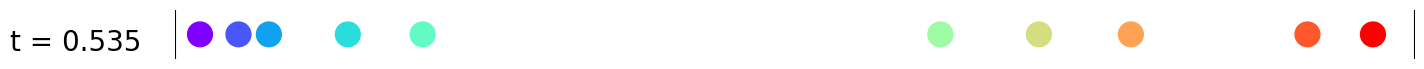}
    \end{minipage}\vfill
    \begin{minipage}{0.78\textwidth}
     \centering
     \includegraphics[width= 1\textwidth]{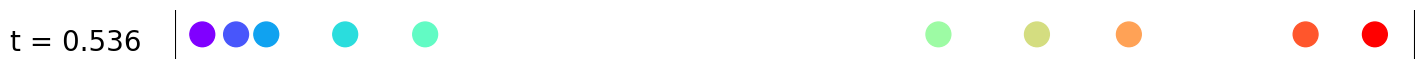}
    \end{minipage}\vfill
    \begin{minipage}{0.78\textwidth}
     \centering
     \includegraphics[width= 1\textwidth]{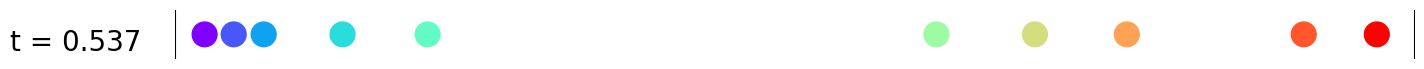}
    \end{minipage}\vfill
    \begin{minipage}{0.78\textwidth}
     \centering
     \includegraphics[width= 1\textwidth]{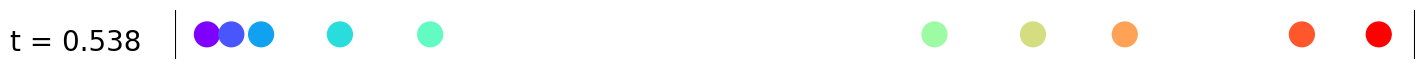}
    \end{minipage}\vfill
    \begin{minipage}{0.78\textwidth}
     \centering
     \includegraphics[width= 1\textwidth]{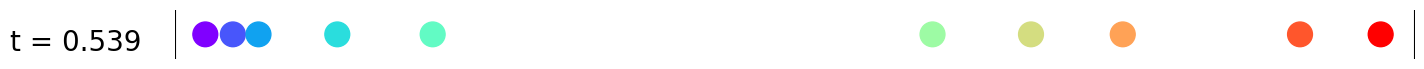}
    \end{minipage}\vfill
    \begin{minipage}{0.78\textwidth}
     \centering
     \includegraphics[width= 1\textwidth]{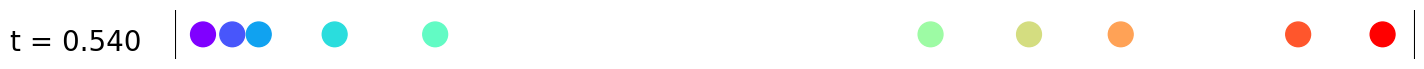}
    \end{minipage}\vfill
    \begin{minipage}{0.78\textwidth}
     \centering
     \includegraphics[width= 1\textwidth]{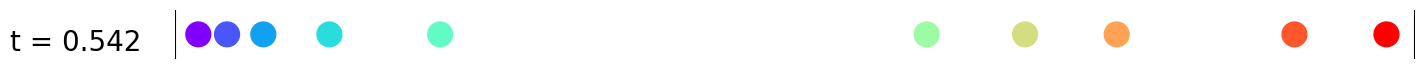}
    \end{minipage}\vfill
\begin{minipage}{0.78\textwidth}
     \centering
     \includegraphics[width= 1\textwidth]{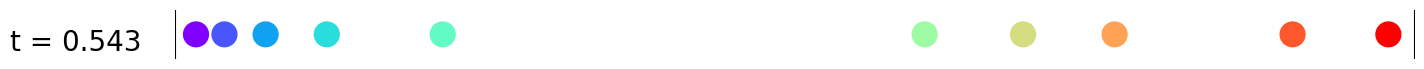}
    \end{minipage}\vfill
    \begin{minipage}{0.78\textwidth}
     \centering
     \includegraphics[width= 1\textwidth]{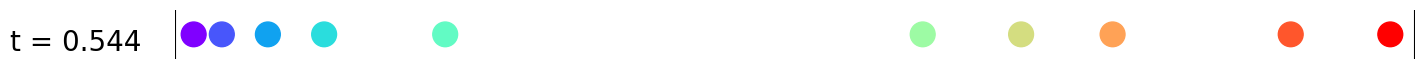}
    \end{minipage}\vfill
    \begin{minipage}{0.78\textwidth}
     \centering
     \includegraphics[width= 1\textwidth]{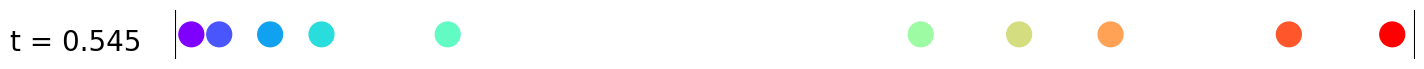}
    \end{minipage}\vfill
    \caption{10 sea floes dataset snapshot}\label{Fig:10nodes_dataset_snapshot}
\end{figure}

\begin{figure}[!htb]
   
    \begin{minipage}{0.98\textwidth}
     \centering
     \includegraphics[width= 1\textwidth]{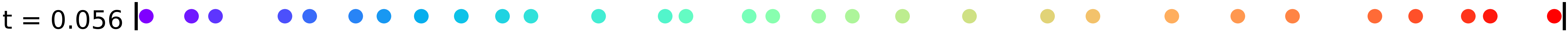}
    \end{minipage}\vfill
    \begin{minipage}{0.98\textwidth}
     \centering
     \includegraphics[width= 1\textwidth]{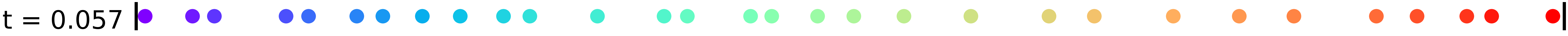}
    \end{minipage}\vfill
    \begin{minipage}{0.98\textwidth}
     \centering
     \includegraphics[width= 1\textwidth]{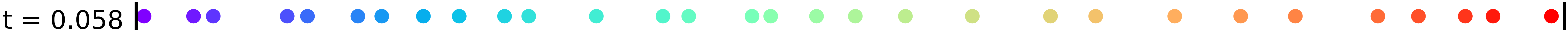}
    \end{minipage}\vfill
    \begin{minipage}{0.98\textwidth}
     \centering
     \includegraphics[width= 1\textwidth]{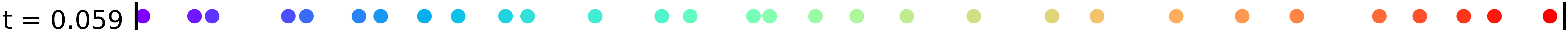}
    \end{minipage}\vfill
    \begin{minipage}{0.98\textwidth}
     \centering
     \includegraphics[width= 1\textwidth]{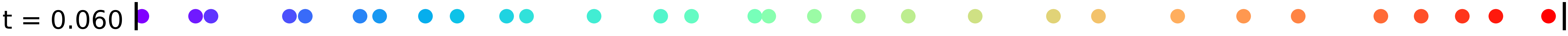}
    \end{minipage}\vfill
    \begin{minipage}{0.98\textwidth}
     \centering
     \includegraphics[width= 1\textwidth]{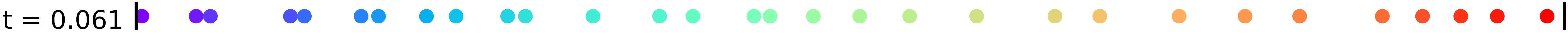}
    \end{minipage}\vfill
    \begin{minipage}{0.98\textwidth}
     \centering
     \includegraphics[width= 1\textwidth]{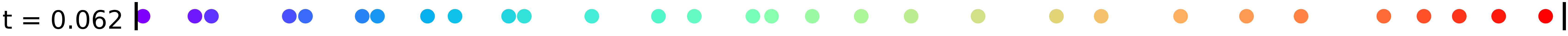}
    \end{minipage}\vfill
    \begin{minipage}{0.98\textwidth}
     \centering
     \includegraphics[width= 1\textwidth]{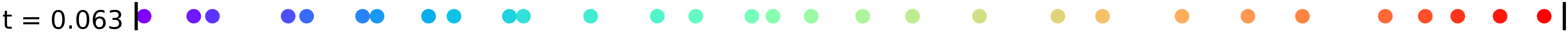}
    \end{minipage}\vfill
    \begin{minipage}{0.98\textwidth}
     \centering
     \includegraphics[width= 1\textwidth]{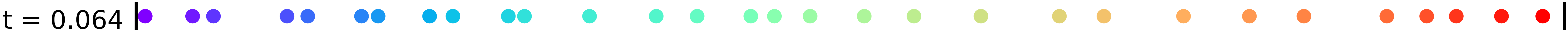}
    \end{minipage}\vfill
    \begin{minipage}{0.98\textwidth}
     \centering
     \includegraphics[width= 1\textwidth]{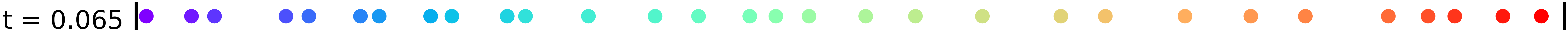}
    \end{minipage}\vfill
\begin{minipage}{0.98\textwidth}
     \centering
     \includegraphics[width= 1\textwidth]{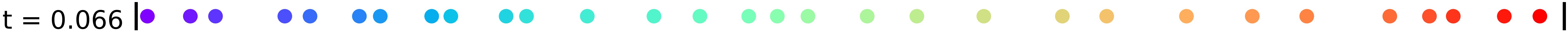}
    \end{minipage}\vfill
    \begin{minipage}{0.98\textwidth}
     \centering
     \includegraphics[width= 1\textwidth]{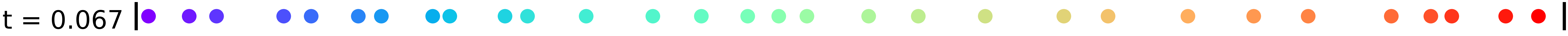}
    \end{minipage}\vfill
    \begin{minipage}{0.98\textwidth}
     \centering
     \includegraphics[width= 1\textwidth]{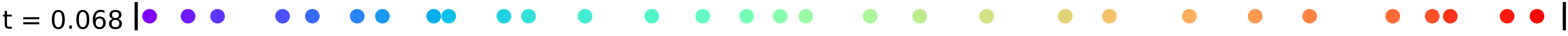}
    \end{minipage}\vfill
    \begin{minipage}{0.98\textwidth}
     \centering
     \includegraphics[width= 1\textwidth]{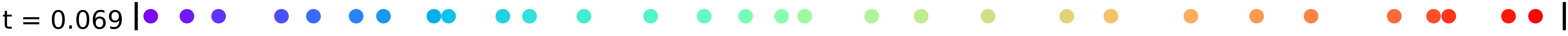}
    \end{minipage}\vfill
    \begin{minipage}{0.98\textwidth}
     \centering
     \includegraphics[width= 1\textwidth]{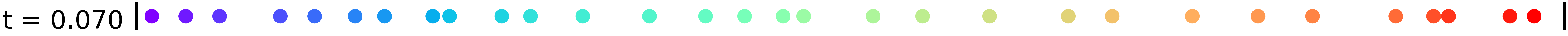}
    \end{minipage}\vfill
    \begin{minipage}{0.98\textwidth}
     \centering
     \includegraphics[width= 1\textwidth]{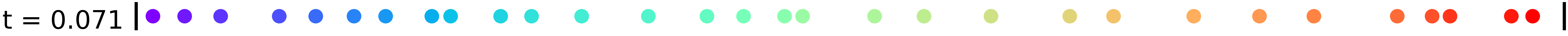}
    \end{minipage}\vfill
    \begin{minipage}{0.98\textwidth}
     \centering
     \includegraphics[width= 1\textwidth]{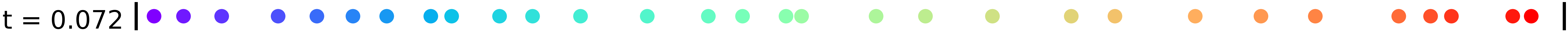}
    \end{minipage}\vfill
    \begin{minipage}{0.98\textwidth}
     \centering
     \includegraphics[width= 1\textwidth]{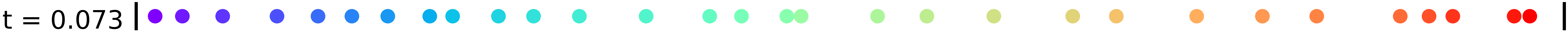}
    \end{minipage}\vfill
    \begin{minipage}{0.98\textwidth}
     \centering
     \includegraphics[width= 1\textwidth]{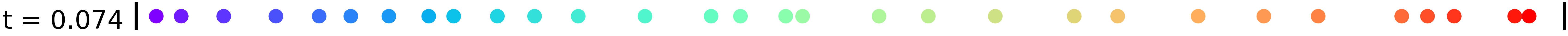}
    \end{minipage}\vfill
    \begin{minipage}{0.98\textwidth}
     \centering
     \includegraphics[width= 1\textwidth]{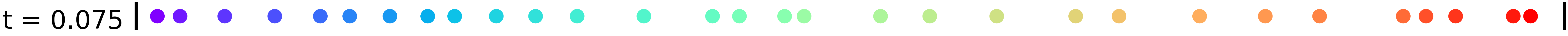}
    \end{minipage}\vfill
\begin{minipage}{0.98\textwidth}
     \centering
     \includegraphics[width= 1\textwidth]{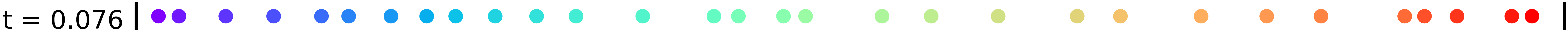}
    \end{minipage}\vfill
    \begin{minipage}{0.98\textwidth}
     \centering
     \includegraphics[width= 1\textwidth]{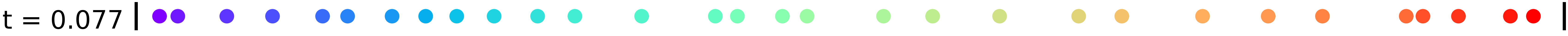}
    \end{minipage}\vfill
    \begin{minipage}{0.98\textwidth}
     \centering
     \includegraphics[width= 1\textwidth]{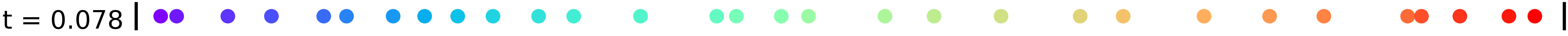}
    \end{minipage}\vfill
    \begin{minipage}{0.98\textwidth}
     \centering
     \includegraphics[width= 1\textwidth]{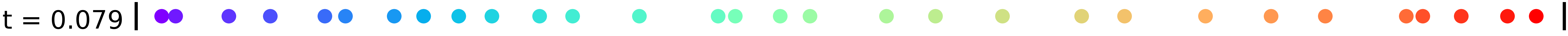}
    \end{minipage}\vfill
    \begin{minipage}{0.98\textwidth}
     \centering
     \includegraphics[width= 1\textwidth]{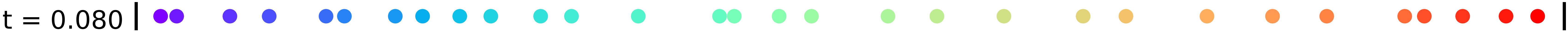}
    \end{minipage}\vfill
    \begin{minipage}{0.98\textwidth}
     \centering
     \includegraphics[width= 1\textwidth]{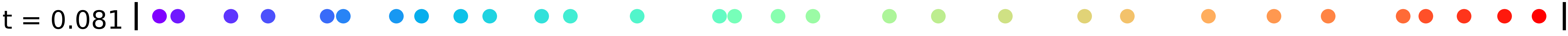}
    \end{minipage}\vfill
    \begin{minipage}{0.98\textwidth}
     \centering
     \includegraphics[width= 1\textwidth]{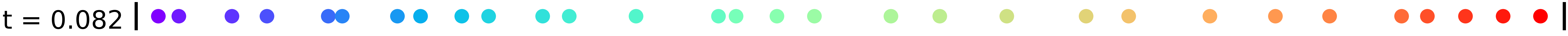}
    \end{minipage}\vfill
    \begin{minipage}{0.98\textwidth}
     \centering
     \includegraphics[width= 1\textwidth]{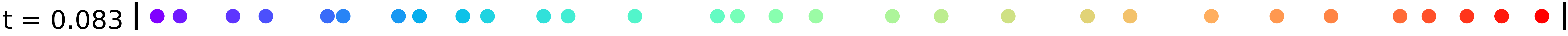}
    \end{minipage}\vfill
    \caption{30 sea floes dataset snapshot}\label{Fig:30nodes_dataset_snapshot}
\end{figure}

The video for dataset can be found at \href{ https://doi.org/10.6084/m9.figshare.28658180}{figshare}\cite{ZHU:data:2025}. This shows the generated dataset follows physical rules.
\clearpage
\subsection{Prediction visualization}

\begin{figure}[!htb]
   \begin{minipage}{0.98\textwidth}
     \centering
     \includegraphics[width= 1\textwidth]{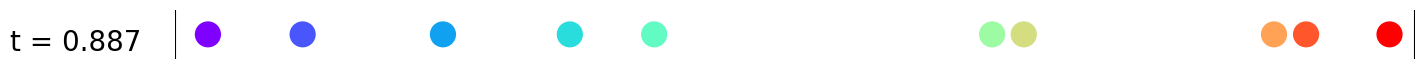}

   \end{minipage}\vfill
   \begin{minipage}{0.98\textwidth}
     \centering
     \includegraphics[width= 1\textwidth]{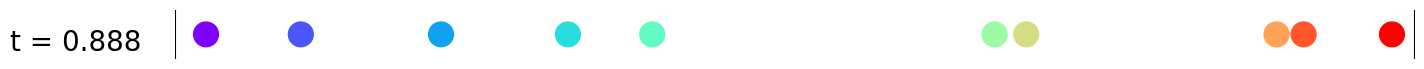}
  
   \end{minipage}\vfill
   \begin{minipage}{0.98\textwidth}
     \centering
     \includegraphics[width= 1\textwidth]{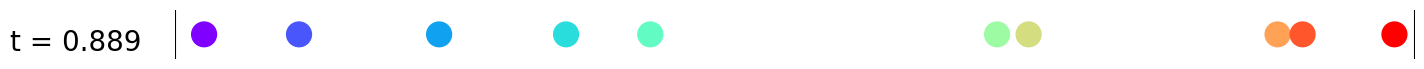}

   \end{minipage}\vfill
   \begin{minipage}{0.98\textwidth}
     \centering
     \includegraphics[width= 1\textwidth]{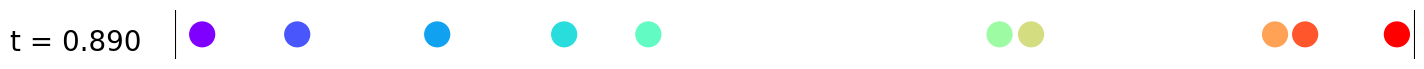}
     
   \end{minipage}\vfill
   \begin{minipage}{0.98\textwidth}
     \centering
     \includegraphics[width= 1\textwidth]{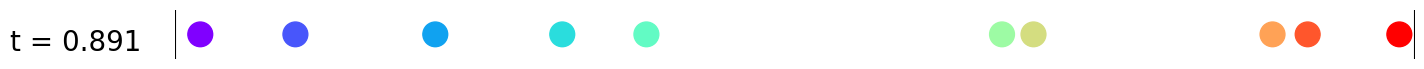}

   \end{minipage}\vfill
   \begin{minipage}{0.98\textwidth}
     \centering
     \includegraphics[width= 1\textwidth]{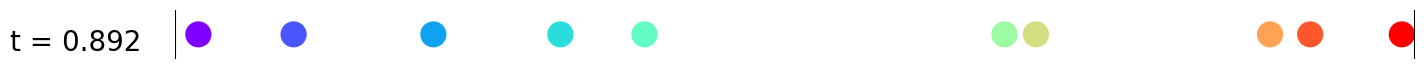}

   \end{minipage}\vfill
   \begin{minipage}{0.98\textwidth}
     \centering
     \includegraphics[width= 1\textwidth]{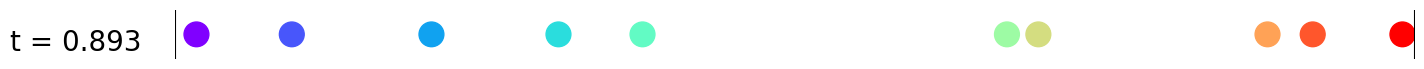}

   \end{minipage}\vfill
   \begin{minipage}{0.98\textwidth}
     \centering
     \includegraphics[width= 1\textwidth]{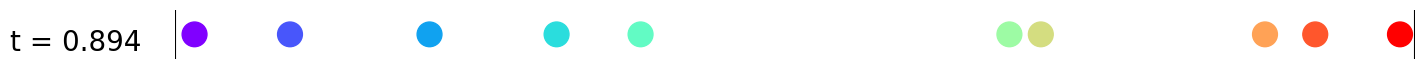}
    
   \end{minipage}\vfill\begin{minipage}{0.98\textwidth}
     \centering
     \includegraphics[width= 1\textwidth]{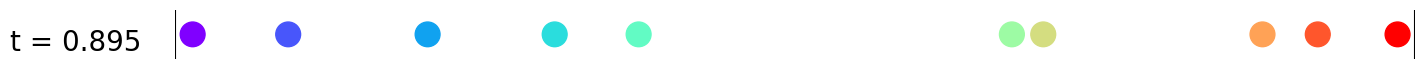}

   \end{minipage}\vfill
   \begin{minipage}{0.98\textwidth}
     \centering
     \includegraphics[width= 1\textwidth]{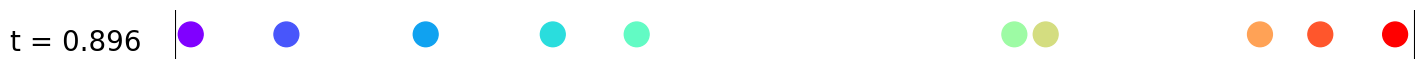}
   \end{minipage}\vfill
   \begin{minipage}{0.98\textwidth}
     \centering
     \includegraphics[width= 1\textwidth]{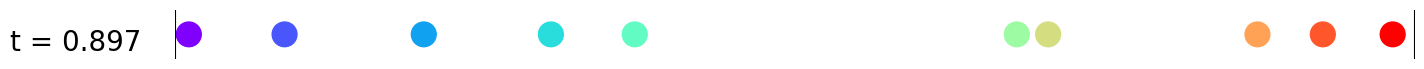}
   \end{minipage}\vfill
   \begin{minipage}{0.98\textwidth}
     \centering
     \includegraphics[width= 1\textwidth]{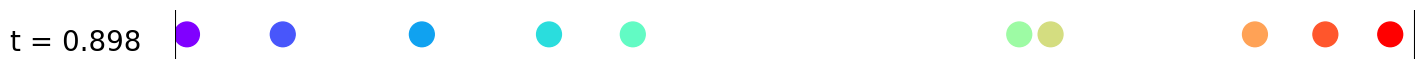}
   \end{minipage}\vfill
   \begin{minipage}{0.98\textwidth}
     \centering
     \includegraphics[width= 1\textwidth]{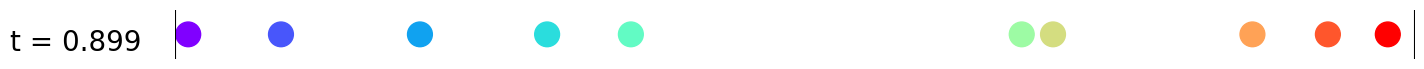}
   \end{minipage}\vfill
   \begin{minipage}{0.98\textwidth}
     \centering
     \includegraphics[width= 1\textwidth]{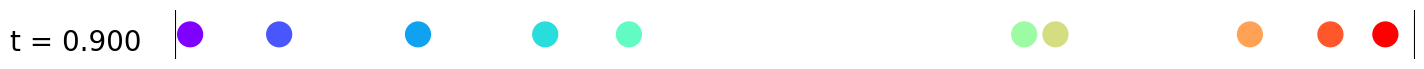}
   \end{minipage}\vfill
   \begin{minipage}{0.98\textwidth}
     \centering
     \includegraphics[width= 1\textwidth]{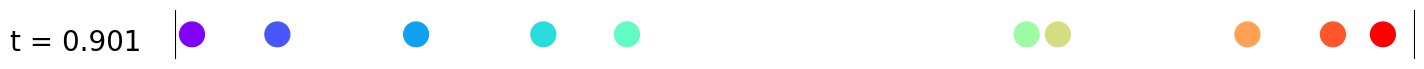}
   \end{minipage}\vfill
   \begin{minipage}{0.98\textwidth}
     \centering
     \includegraphics[width= 1\textwidth]{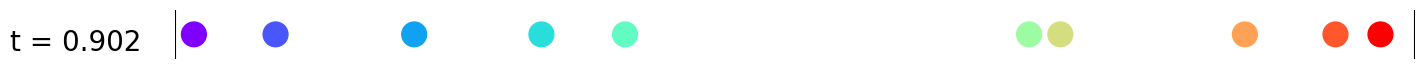}
   \end{minipage}\vfill
   \begin{minipage}{0.98\textwidth}
     \centering
     \includegraphics[width= 1\textwidth]{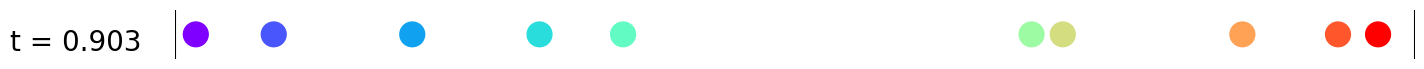}
   \end{minipage}\vfill
   \begin{minipage}{0.98\textwidth}
     \centering
     \includegraphics[width= 1\textwidth]{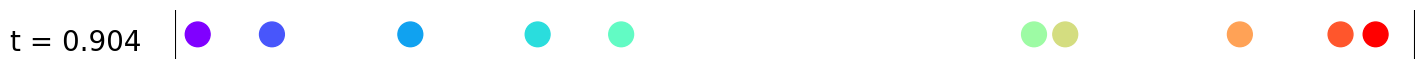}
    
   \end{minipage}\vfill\begin{minipage}{0.98\textwidth}
     \centering
     \includegraphics[width= 1\textwidth]{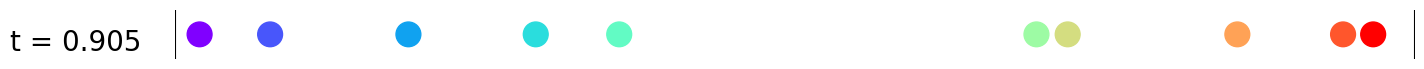}

   \end{minipage}\vfill
   \begin{minipage}{0.98\textwidth}
     \centering
     \includegraphics[width= 1\textwidth]{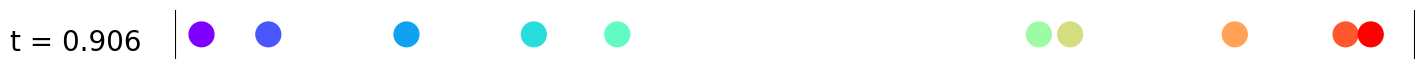}
   \end{minipage}\vfill
   \begin{minipage}{0.98\textwidth}
     \centering
     \includegraphics[width= 1\textwidth]{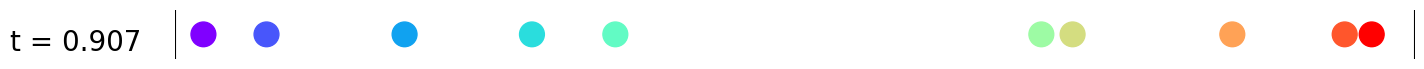}
   \end{minipage}\vfill
   \begin{minipage}{0.98\textwidth}
     \centering
     \includegraphics[width= 1\textwidth]{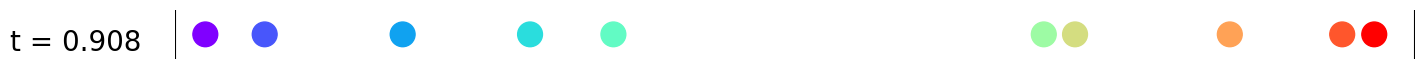}
   \end{minipage}\vfill
   \begin{minipage}{0.98\textwidth}
     \centering
     \includegraphics[width= 1\textwidth]{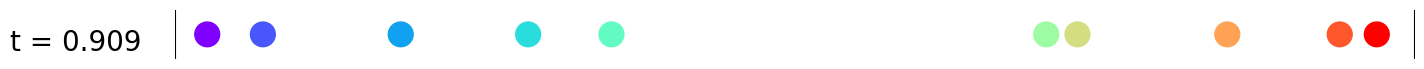}
   \end{minipage}\vfill
   \caption{10 sea floes prediction snapshot}\label{Fig:10nodes_snapshot}
\end{figure}

\begin{figure}[!htb]
   \begin{minipage}{0.98\textwidth}
     \centering
     \includegraphics[width= 1\textwidth]{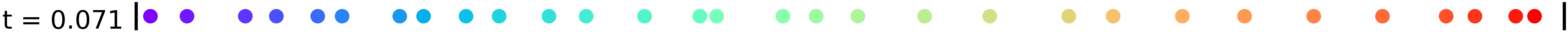}
    \end{minipage}\vfill
\begin{minipage}{0.98\textwidth}
     \centering
     \includegraphics[width= 1\textwidth]{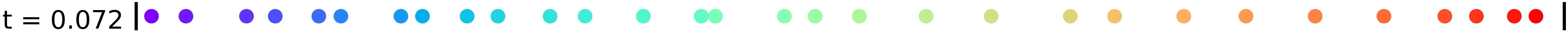}
    \end{minipage}\vfill
    \begin{minipage}{0.98\textwidth}
     \centering
     \includegraphics[width= 1\textwidth]{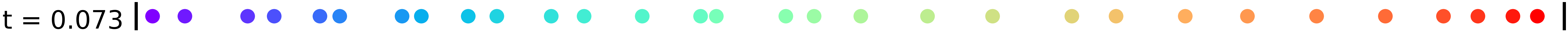}
    \end{minipage}\vfill
    \begin{minipage}{0.98\textwidth}
     \centering
     \includegraphics[width= 1\textwidth]{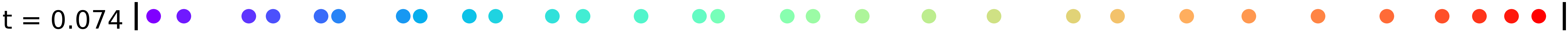}
    \end{minipage}\vfill
    \begin{minipage}{0.98\textwidth}
     \centering
     \includegraphics[width= 1\textwidth]{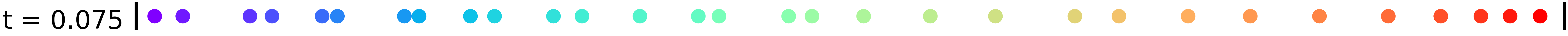}
    \end{minipage}\vfill
    \begin{minipage}{0.98\textwidth}
     \centering
     \includegraphics[width= 1\textwidth]{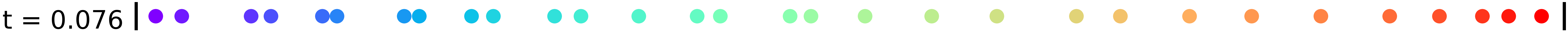}
    \end{minipage}\vfill
    \begin{minipage}{0.98\textwidth}
     \centering
     \includegraphics[width= 1\textwidth]{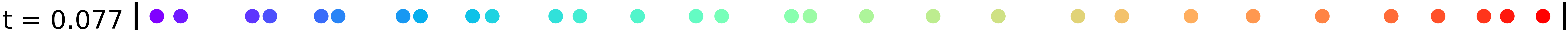}
    \end{minipage}\vfill
    \begin{minipage}{0.98\textwidth}
     \centering
     \includegraphics[width= 1\textwidth]{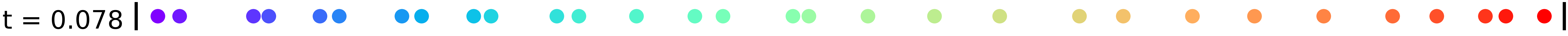}
    \end{minipage}\vfill
    \begin{minipage}{0.98\textwidth}
     \centering
     \includegraphics[width= 1\textwidth]{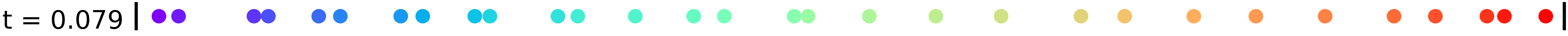}
    \end{minipage}\vfill
    \begin{minipage}{0.98\textwidth}
     \centering
     \includegraphics[width= 1\textwidth]{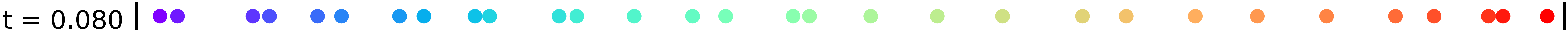}
    \end{minipage}\vfill
    \begin{minipage}{0.98\textwidth}
     \centering
     \includegraphics[width= 1\textwidth]{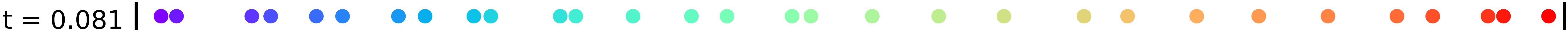}
    \end{minipage}\vfill
\begin{minipage}{0.98\textwidth}
     \centering
     \includegraphics[width= 1\textwidth]{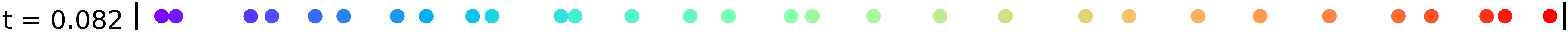}
    \end{minipage}\vfill
    \begin{minipage}{0.98\textwidth}
     \centering
     \includegraphics[width= 1\textwidth]{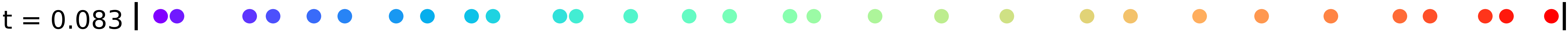}
    \end{minipage}\vfill
    \begin{minipage}{0.98\textwidth}
     \centering
     \includegraphics[width= 1\textwidth]{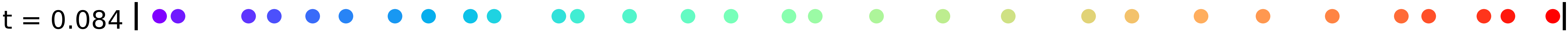}
    \end{minipage}\vfill
    \begin{minipage}{0.98\textwidth}
     \centering
     \includegraphics[width= 1\textwidth]{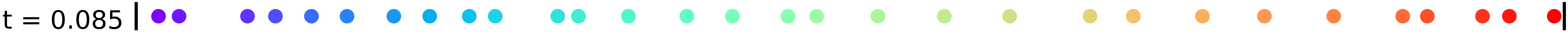}
    \end{minipage}\vfill
    \begin{minipage}{0.98\textwidth}
     \centering
     \includegraphics[width= 1\textwidth]{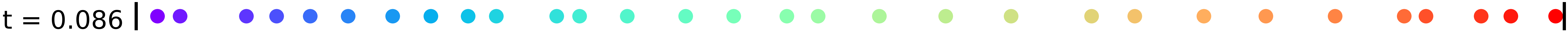}
    \end{minipage}\vfill
    \begin{minipage}{0.98\textwidth}
     \centering
     \includegraphics[width= 1\textwidth]{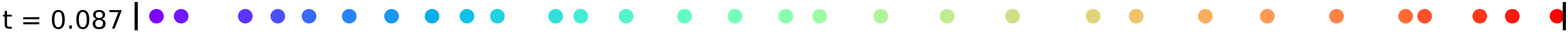}
    \end{minipage}\vfill
    \begin{minipage}{0.98\textwidth}
     \centering
     \includegraphics[width= 1\textwidth]{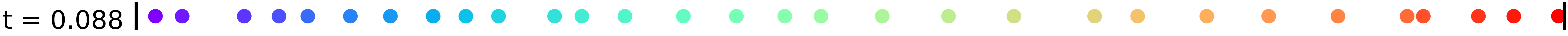}
    \end{minipage}\vfill
    \begin{minipage}{0.98\textwidth}
     \centering
     \includegraphics[width= 1\textwidth]{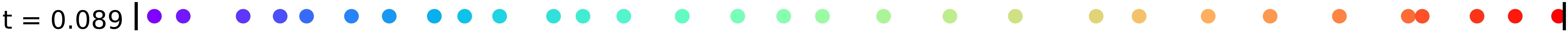}
    \end{minipage}\vfill
    \begin{minipage}{0.98\textwidth}
     \centering
     \includegraphics[width= 1\textwidth]{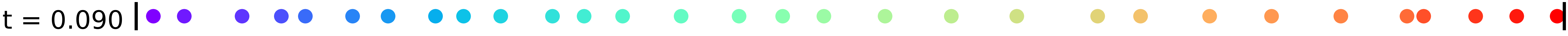}
    \end{minipage}\vfill
    \begin{minipage}{0.98\textwidth}
     \centering
     \includegraphics[width= 1\textwidth]{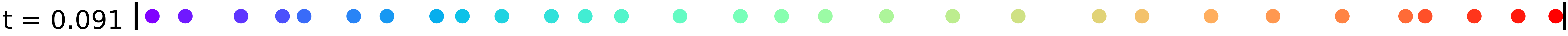}
    \end{minipage}\vfill
\begin{minipage}{0.98\textwidth}
     \centering
     \includegraphics[width= 1\textwidth]{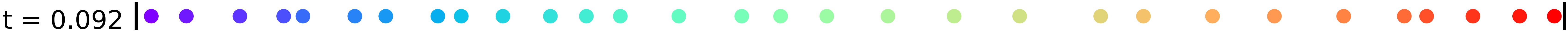}
    \end{minipage}\vfill
    \begin{minipage}{0.98\textwidth}
     \centering
     \includegraphics[width= 1\textwidth]{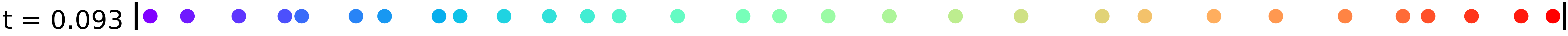}
    \end{minipage}\vfill
    \begin{minipage}{0.98\textwidth}
     \centering
     \includegraphics[width= 1\textwidth]{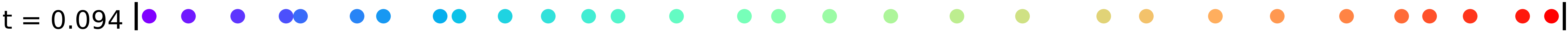}
    \end{minipage}\vfill
    \begin{minipage}{0.98\textwidth}
     \centering
     \includegraphics[width= 1\textwidth]{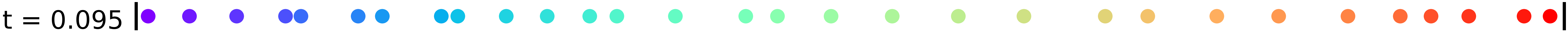}
    \end{minipage}\vfill
    \begin{minipage}{0.98\textwidth}
     \centering
     \includegraphics[width= 1\textwidth]{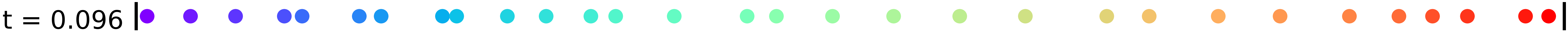}
    \end{minipage}\vfill
    \begin{minipage}{0.98\textwidth}
     \centering
     \includegraphics[width= 1\textwidth]{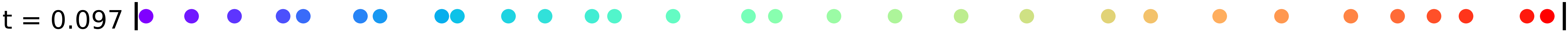}
    \end{minipage}\vfill
    \begin{minipage}{0.98\textwidth}
     \centering
     \includegraphics[width= 1\textwidth]{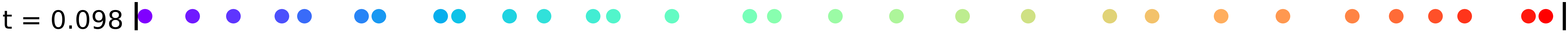}
    \end{minipage}\vfill
    \caption{30 sea floes prediction snapshot}\label{Fig:30nodes_snapshot}
\end{figure}

Figure \ref{Fig:30nodes_snapshot} provides a qualitative snapshot of the predicted dynamics in the 30-floe experiment and illustrates two representative interaction modes captured by the model: floe–floe collisions and floe–boundary collisions. For floe–floe interaction, the first two floes on the left demonstrate a clear approach–contact–separation sequence. Between  $t=0.071$ and  $t=0.082$, the two floes move closer as their distance decreases, and they make contact at $t=0.082$. Immediately after contact, the subsequent frames show that the floes rebound and move in opposite directions, consistent with physical laws. For the floe and boundary interaction, it is visible for the first floe on the right. From $t=0.071$ to  $t=0.085$, this floe translates toward the right boundary, reaches contact at $t=0.085$, and then rebounds, reversing direction and moving away from the boundary thereafter. Together, these sequences demonstrate that the model produces physically plausible interaction patterns across both internal collisions and boundary interactions, supporting the qualitative validity of the learned collision dynamics in addition to the quantitative performance metrics reported.

The videos for prediction can be found at \href{ https://doi.org/10.6084/m9.figshare.28658180}{figshare}\cite{ZHU:data:2025}. These visualizations vividly illustrate the interactions modeled within the proposed model in different node size settings, effectively capturing various collision dynamics. Blue nodes represent individual sea ice floes, while black lines delineate the boundaries of the simulation area. This visualization demonstrates the model's capability to accurately simulate different scenarios: the collision between two floes, a single floe's interaction with the boundary, and cases where no collisions exist. Each scenario is distinctly represented, showcasing the model's comprehensive and precise handling of physical interactions in the dynamic environment. This accuracy substantiates that the proposed model's predictions are visually feasible and do not contravene any physical laws.

\clearpage
\subsection{Generalization visualization}

\begin{figure}[!htb]
   \begin{minipage}{0.98\textwidth}
     \centering
     \includegraphics[width= 1\textwidth]{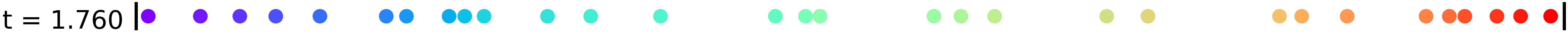}
    \end{minipage}\vfill
\begin{minipage}{0.98\textwidth}
     \centering
     \includegraphics[width= 1\textwidth]{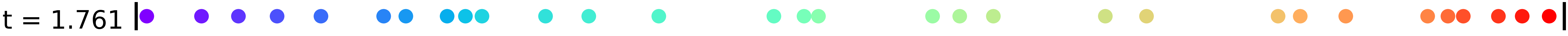}
    \end{minipage}\vfill
    \begin{minipage}{0.98\textwidth}
     \centering
     \includegraphics[width= 1\textwidth]{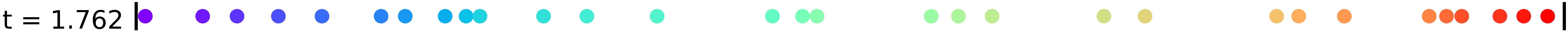}
    \end{minipage}\vfill
    \begin{minipage}{0.98\textwidth}
     \centering
     \includegraphics[width= 1\textwidth]{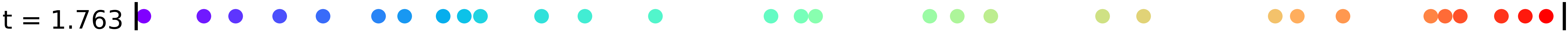}
    \end{minipage}\vfill
    \begin{minipage}{0.98\textwidth}
     \centering
     \includegraphics[width= 1\textwidth]{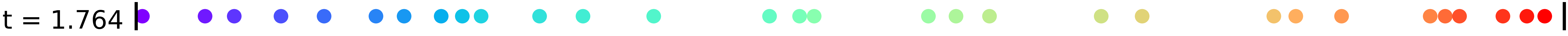}
    \end{minipage}\vfill
    \begin{minipage}{0.98\textwidth}
     \centering
     \includegraphics[width= 1\textwidth]{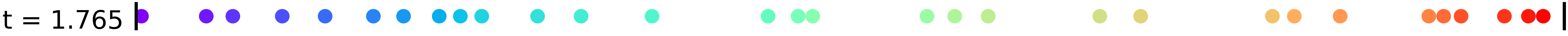}
    \end{minipage}\vfill
    \begin{minipage}{0.98\textwidth}
     \centering
     \includegraphics[width= 1\textwidth]{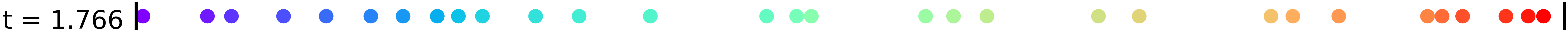}
    \end{minipage}\vfill
    \begin{minipage}{0.98\textwidth}
     \centering
     \includegraphics[width= 1\textwidth]{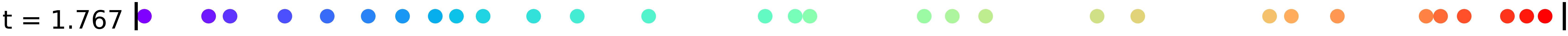}
    \end{minipage}\vfill
    \begin{minipage}{0.98\textwidth}
     \centering
     \includegraphics[width= 1\textwidth]{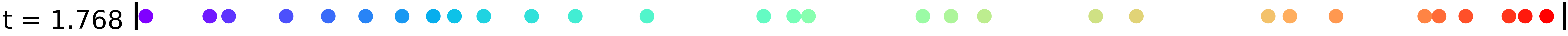}
    \end{minipage}\vfill
    \begin{minipage}{0.98\textwidth}
     \centering
     \includegraphics[width= 1\textwidth]{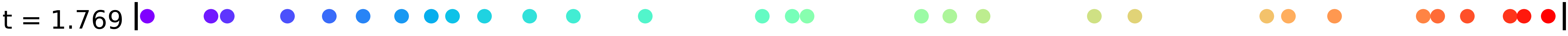}
    \end{minipage}\vfill
    \begin{minipage}{0.98\textwidth}
     \centering
     \includegraphics[width= 1\textwidth]{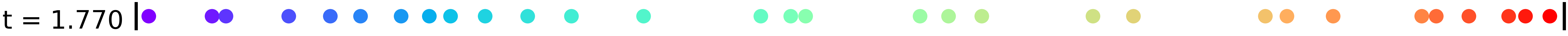}
    \end{minipage}\vfill
\begin{minipage}{0.98\textwidth}
     \centering
     \includegraphics[width= 1\textwidth]{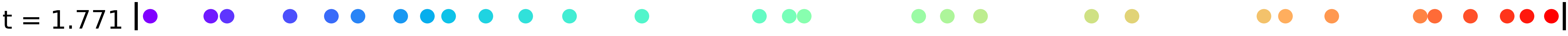}
    \end{minipage}\vfill
    \begin{minipage}{0.98\textwidth}
     \centering
     \includegraphics[width= 1\textwidth]{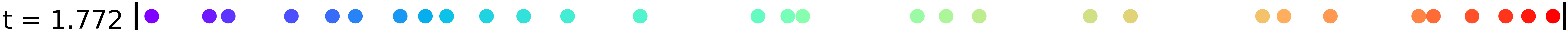}
    \end{minipage}\vfill
    \begin{minipage}{0.98\textwidth}
     \centering
     \includegraphics[width= 1\textwidth]{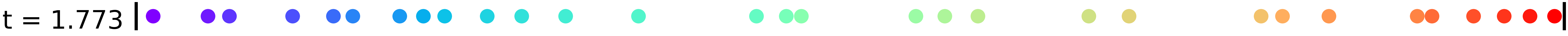}
    \end{minipage}\vfill
    \begin{minipage}{0.98\textwidth}
     \centering
     \includegraphics[width= 1\textwidth]{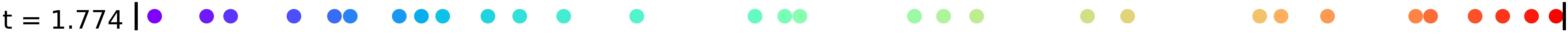}
    \end{minipage}\vfill
    \begin{minipage}{0.98\textwidth}
     \centering
     \includegraphics[width= 1\textwidth]{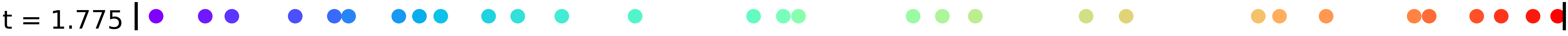}
    \end{minipage}\vfill
    \begin{minipage}{0.98\textwidth}
     \centering
     \includegraphics[width= 1\textwidth]{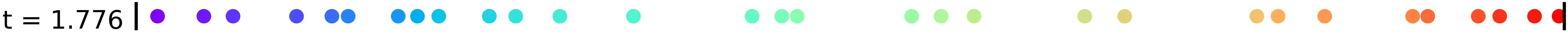}
    \end{minipage}\vfill
    \begin{minipage}{0.98\textwidth}
     \centering
     \includegraphics[width= 1\textwidth]{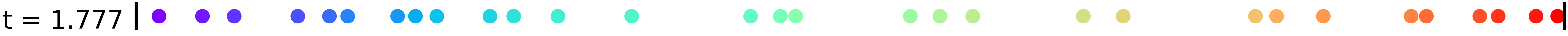}
    \end{minipage}\vfill
    \begin{minipage}{0.98\textwidth}
     \centering
     \includegraphics[width= 1\textwidth]{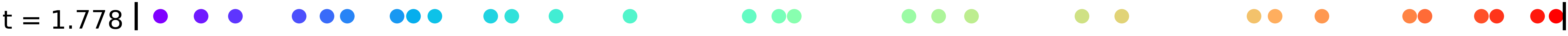}
    \end{minipage}\vfill
    \begin{minipage}{0.98\textwidth}
     \centering
     \includegraphics[width= 1\textwidth]{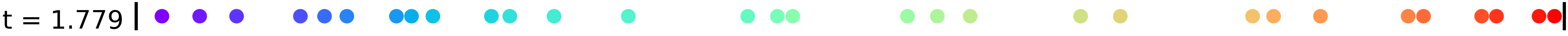}
    \end{minipage}\vfill
    \begin{minipage}{0.98\textwidth}
     \centering
     \includegraphics[width= 1\textwidth]{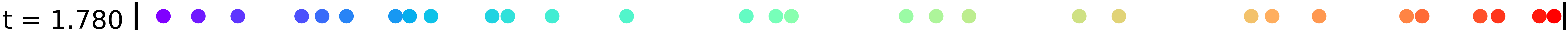}
    \end{minipage}\vfill
\begin{minipage}{0.98\textwidth}
     \centering
     \includegraphics[width= 1\textwidth]{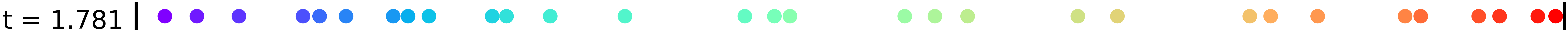}
    \end{minipage}\vfill
    \begin{minipage}{0.98\textwidth}
     \centering
     \includegraphics[width= 1\textwidth]{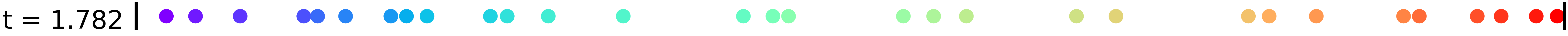}
    \end{minipage}\vfill
    \begin{minipage}{0.98\textwidth}
     \centering
     \includegraphics[width= 1\textwidth]{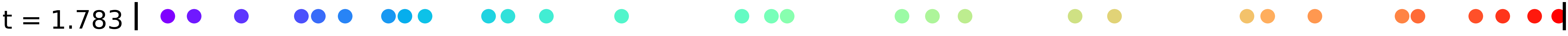}
    \end{minipage}\vfill
    \begin{minipage}{0.98\textwidth}
     \centering
     \includegraphics[width= 1\textwidth]{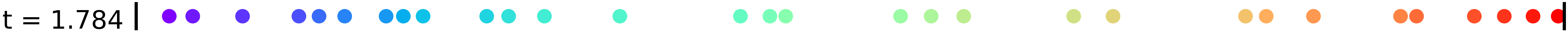}
    \end{minipage}\vfill
    \begin{minipage}{0.98\textwidth}
     \centering
     \includegraphics[width= 1\textwidth]{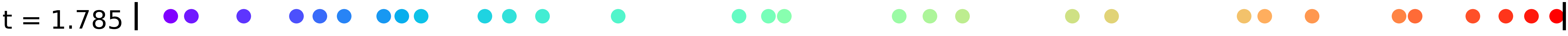}
    \end{minipage}\vfill
    \begin{minipage}{0.98\textwidth}
     \centering
     \includegraphics[width= 1\textwidth]{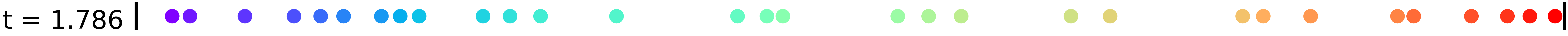}
    \end{minipage}\vfill
    \begin{minipage}{0.98\textwidth}
     \centering
     \includegraphics[width= 1\textwidth]{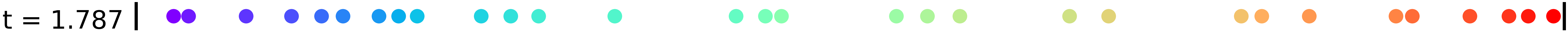}
    \end{minipage}\vfill
    \begin{minipage}{0.98\textwidth}
     \centering
     \includegraphics[width= 1\textwidth]{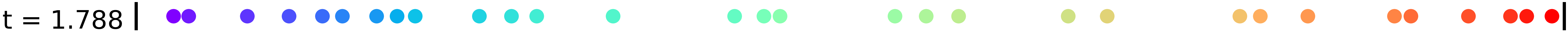}
    \end{minipage}\vfill
    \caption{30 sea floes generalization snapshot}\label{Fig:30nodes_generalization_snapshot}
\end{figure}
\begin{figure}[!htb]
   \begin{minipage}{0.98\textwidth}
     \centering
     \includegraphics[width= 1\textwidth]{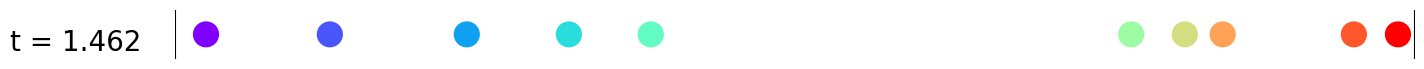}
    \end{minipage}\vfill
\begin{minipage}{0.98\textwidth}
     \centering
     \includegraphics[width= 1\textwidth]{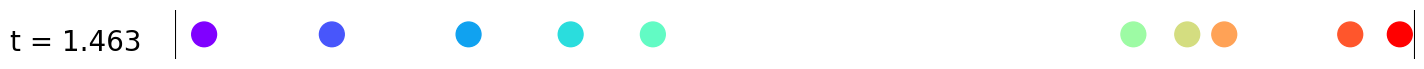}
    \end{minipage}\vfill
    \begin{minipage}{0.98\textwidth}
     \centering
     \includegraphics[width= 1\textwidth]{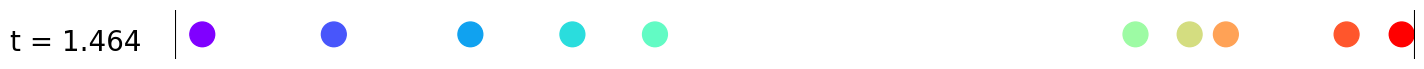}
    \end{minipage}\vfill
    \begin{minipage}{0.98\textwidth}
     \centering
     \includegraphics[width= 1\textwidth]{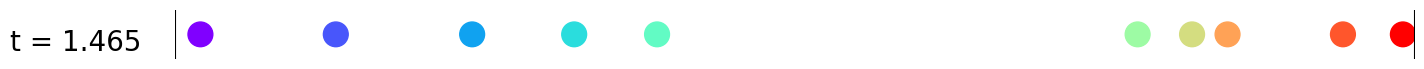}
    \end{minipage}\vfill
    \begin{minipage}{0.98\textwidth}
     \centering
     \includegraphics[width= 1\textwidth]{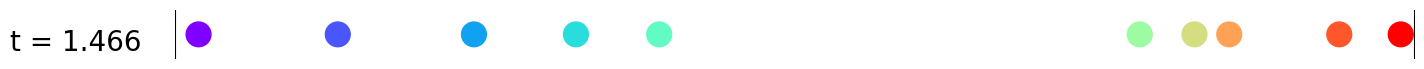}
    \end{minipage}\vfill
    \begin{minipage}{0.98\textwidth}
     \centering
     \includegraphics[width= 1\textwidth]{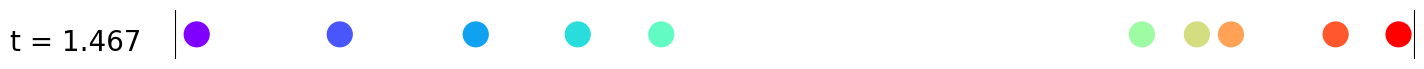}
    \end{minipage}\vfill
    \begin{minipage}{0.98\textwidth}
     \centering
     \includegraphics[width= 1\textwidth]{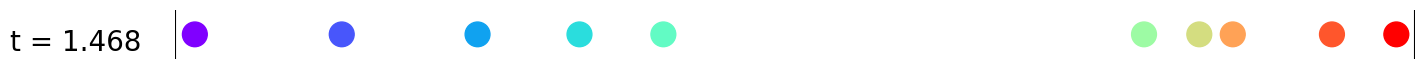}
    \end{minipage}\vfill
    \begin{minipage}{0.98\textwidth}
     \centering
     \includegraphics[width= 1\textwidth]{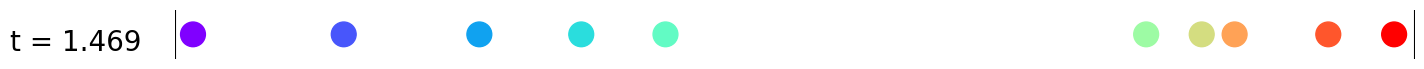}
    \end{minipage}\vfill
    \begin{minipage}{0.98\textwidth}
     \centering
     \includegraphics[width= 1\textwidth]{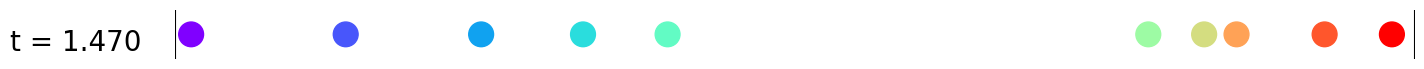}
    \end{minipage}\vfill
    \begin{minipage}{0.98\textwidth}
     \centering
     \includegraphics[width= 1\textwidth]{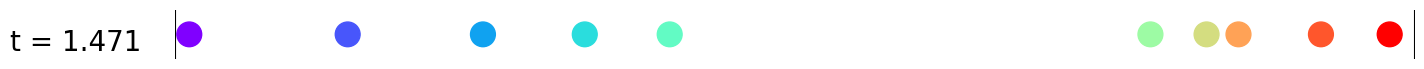}
    \end{minipage}\vfill
    \begin{minipage}{0.98\textwidth}
     \centering
     \includegraphics[width= 1\textwidth]{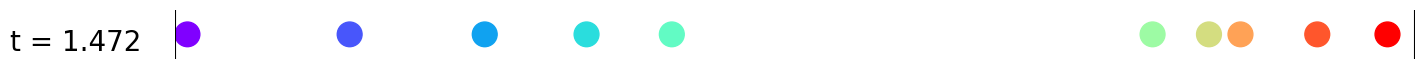}
    \end{minipage}\vfill
\begin{minipage}{0.98\textwidth}
     \centering
     \includegraphics[width= 1\textwidth]{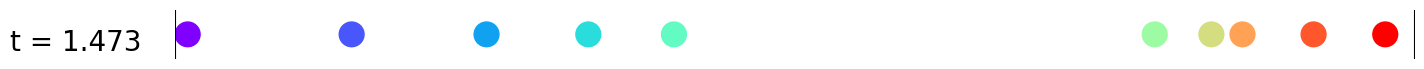}
    \end{minipage}\vfill
    \begin{minipage}{0.98\textwidth}
     \centering
     \includegraphics[width= 1\textwidth]{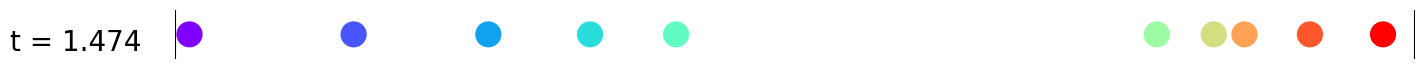}
    \end{minipage}\vfill
    \begin{minipage}{0.98\textwidth}
     \centering
     \includegraphics[width= 1\textwidth]{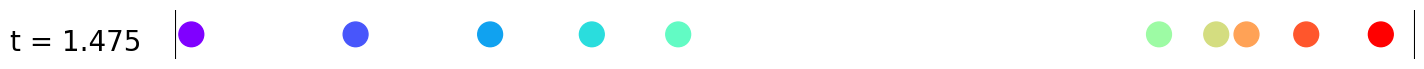}
    \end{minipage}\vfill
    \begin{minipage}{0.98\textwidth}
     \centering
     \includegraphics[width= 1\textwidth]{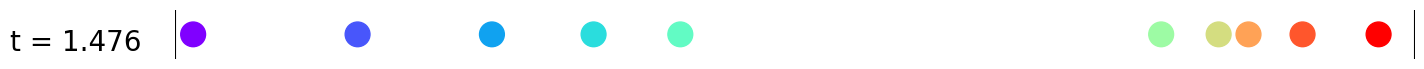}
    \end{minipage}\vfill
    \begin{minipage}{0.98\textwidth}
     \centering
     \includegraphics[width= 1\textwidth]{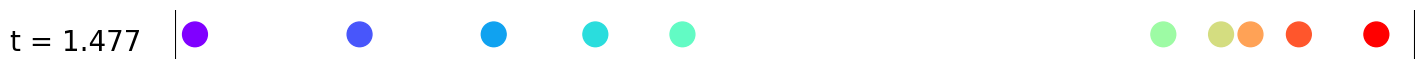}
    \end{minipage}\vfill
    \begin{minipage}{0.98\textwidth}
     \centering
     \includegraphics[width= 1\textwidth]{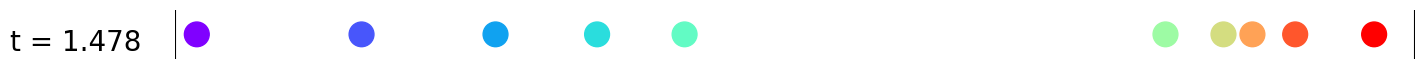}
    \end{minipage}\vfill
    \begin{minipage}{0.98\textwidth}
     \centering
     \includegraphics[width= 1\textwidth]{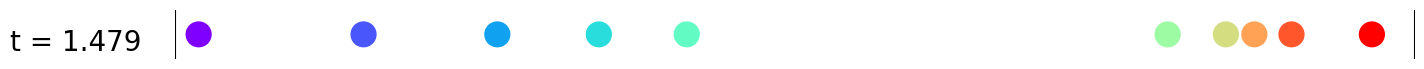}
    \end{minipage}\vfill
    \begin{minipage}{0.98\textwidth}
     \centering
     \includegraphics[width= 1\textwidth]{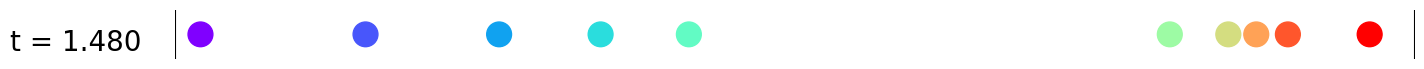}
    \end{minipage}\vfill
    \begin{minipage}{0.98\textwidth}
     \centering
     \includegraphics[width= 1\textwidth]{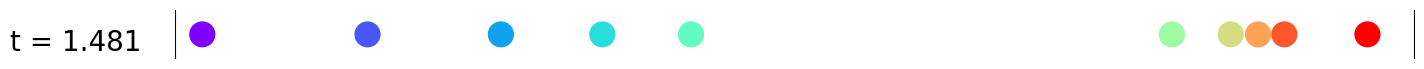}
    \end{minipage}\vfill
    \begin{minipage}{0.98\textwidth}
     \centering
     \includegraphics[width= 1\textwidth]{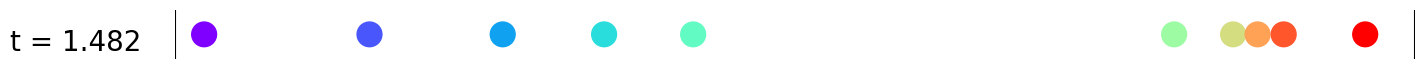}
    \end{minipage}\vfill
\begin{minipage}{0.98\textwidth}
     \centering
     \includegraphics[width= 1\textwidth]{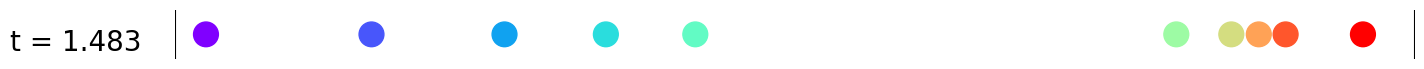}
    \end{minipage}\vfill
    \begin{minipage}{0.98\textwidth}
     \centering
     \includegraphics[width= 1\textwidth]{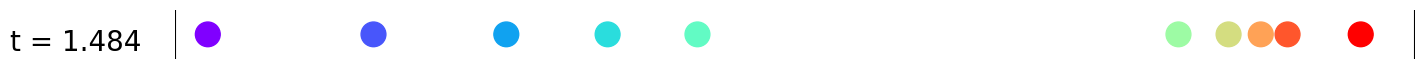}
    \end{minipage}\vfill
    
    \caption{10 sea floes generalization snapshot}\label{Fig:10nodes_generalization_snapshot}
\end{figure}

The video for generalization can be found at \href{ https://doi.org/10.6084/m9.figshare.28658180}{figshare} \cite{ZHU:data:2025}. The visualization for generalization outside the ground truth data shows the proposed model's capability of generalization outside the original time domain.

\newpage

\end{appendices}


\end{document}